%% file: 0main.tex
\pdfoutput=1

\documentclass[11pt]{article}

\usepackage[]{emnlp2021}
\usepackage{times}
\usepackage{latexsym}

\usepackage[T1]{fontenc}

\usepackage[utf8]{inputenc}

\usepackage{microtype}
\usepackage{microtype}
\usepackage{amsmath,amsfonts,amssymb}
\usepackage{bm}
\usepackage{booktabs}
\usepackage{multirow}
\usepackage{graphicx}
\usepackage{times}
\usepackage{latexsym}

\usepackage{graphicx}
\usepackage{caption}
\usepackage{subcaption}
\usepackage[inline]{enumitem}
\usepackage{varwidth}
\usepackage{todonotes}
\usepackage[framemethod=TikZ]{mdframed}
\usepackage{amsfonts}
\usepackage{array,booktabs,arydshln,xcolor}
\usepackage{graphics}
 \usepackage[most]{tcolorbox}
\usepackage{lipsum}
\usepackage{pifont}
\usepackage{hyperref}
\usepackage{makecell}
\usepackage[super]{nth}
\usepackage{wasysym}
\usepackage{float}
\usepackage{resizegather}
\usepackage{adjustbox}
\usepackage{booktabs}

\DeclareMathOperator{\E}{\mathbb{E}}

\def\argmax{\operatornamewithlimits{arg\,max}}

\title{Identifying Morality Frames in Political Tweets using Relational Learning}

\author{Shamik Roy, Maria Leonor Pacheco, Dan Goldwasser \\
        Department of Computer Science\\
        Purdue University\\
        West Lafayette, IN, USA\\
        \texttt{\{roy98, pachecog, dgoldwas\}@purdue.edu}}

\begin{document}
\maketitle
\begin{abstract}
Extracting moral sentiment from text is a vital component in understanding public opinion, social movements, and policy decisions. The Moral Foundation Theory identifies five moral foundations, each associated with a positive and negative polarity. However, moral sentiment is often motivated by its targets, which can correspond to individuals or collective entities. In this paper, we introduce morality frames, a representation framework for organizing moral attitudes directed at different entities, and come up with a novel and high-quality annotated dataset of tweets written by US politicians. Then, we propose a relational learning model to predict moral attitudes towards entities and moral foundations jointly. We do qualitative and quantitative evaluations, showing that moral sentiment towards entities differs highly across political ideologies.
\end{abstract}

\input{1introduction.tex}

\input{2related.tex}
\input{3data.tex}

\input{4model.tex}

\input{5eval.tex}

\input{6qualitative}

\input{7summary.tex}

\clearpage
\input{8ethical.tex}

\bibliographystyle{acl_natbib}
\bibliography{anthology,acl2021}


\input{appendix.tex}

\end{document}

%% file: 1introduction.tex
\section{Introduction}\label{sec:introduction}

Morality is a set of principles to distinguish between right and wrong. 
Shared moral values form the social and cultural norms that unite social groups~\cite{dehghani2016purity}. \textit{Moral Foundations Theory} (MFT)~\cite{haidt2004intuitive,haidt2007morality} provides a theoretical framework for analyzing different expressions of moral values. The theory suggests that there are at least five basic moral values, emerging from evolutionary, social, and cultural origins. These are referred to as Moral Foundations (MFs), each with a positive and a negative polarity, and include \textit{Care/Harm, Fairness/Cheating, Loyalty/Betrayal, Authority/Subversion}, and \textit{Purity/Degradation} (Table~\ref{tab:moral-foundation-roles} provides details).  Identifying MF in text is a relatively new challenge and past work has relied on lexical resources such as the Moral Foundation Dictionary~\cite{DVN/SJTRBI_2009,fulgoni-etal-2016-empirical,xie2020text} and annotated data~\cite{johnson-goldwasser-2018-classification,lin2018acquiring,hoover2020moral}.

Social and political science studies have repeatedly shown the correlation between ideological and political stances and moral foundation preferences~\cite{DVN/SJTRBI_2009,wolsko2016red,amin2017association}. For example, \citeauthor{DVN/SJTRBI_2009}, \citeyear{DVN/SJTRBI_2009} captures the correlation between political ideology and moral foundation usage, showing that Liberals have a preference for Care/Harm and Fairness/Cheating while Conservatives use all five. 
%
%
 Our main intuition in this paper is that even when different groups use the same MF, the moral sentiment would be directed towards different targets. To clarify, consider the following tweets discussing the Affordable Healthcare Act (ACA, Obamacare).

\begin{tcbraster}[raster columns=1,raster equal height=rows,raster valign=top, size=small]
\begin{tcolorbox}[colback=red!15!white,colframe=red!75!black,nobeforeafter, title={}]
\small{{
\textcolor{purple}{$[$}@SenThadCochran and  I\textcolor{purple}{$]_{CARING}$} are working to protect\\ \textcolor{blue}{$[$}MS small businesses\textcolor{blue}{$]_{CARE-FOR}$} from more expensive\\ \textcolor{red}{$[$}\#Obamacare mandates\textcolor{red}{$]_{HARMING}$}.
}}
\end{tcolorbox}  
\begin{tcolorbox}[colback=blue!15!white,colframe=blue!75!black,nobeforeafter, title={}]
\small{{ 
\textcolor{purple}{$[$}The ACA\textcolor{purple}{$]_{CARING}$} was a life saver for the more than \textcolor{blue}{$[$}130 million Americans\textcolor{blue}{$]_{CARE-FOR}$} with a preexisting condition – including covid now. \textcolor{red}{$[$}Republicans\textcolor{red}{$]_{HARMING}$} want to take us back to coverage denials.
}}
\end{tcolorbox}  
\end{tcbraster}
\vspace{2pt}
%
While both tweets use the Care/Harm MF, in the top tweet (Conservative) the ACA is described as causing Harm, while in the bottom (Liberal), the ACA is described as providing the needed Care. 

Our main contribution in this paper is to introduce \textit{morality frames}, a representation framework for organizing moral attitudes directed at different targets, by decomposing the moral foundations into structured frames, each associated with a predicate (a specific MF) and typed roles. For example, the morality frame for Care/Harm is associated with three typed roles: entity providing care, entity needing the care, and entity causing harm. We focus on analyzing political tweets, each describing an eliciting situation that evokes the moral sentiment, and map the text to a MF, and the entities appearing in it to typed roles. Given tweets by different ideological groups discussing the same real-world situation, morality frames can provide the means to explain and compare the attitudes of the two groups. We build on the MF dataset by~\citeauthor{johnson-goldwasser-2018-classification}, \citeyear{johnson-goldwasser-2018-classification} consisting of political tweets, and annotate each tweet with MF roles for its entities.  

Identifying moral roles from text in our setting requires inferences based on \textit{political knowledge}, mapping between the author's perspectives and the judgements appearing in the text. For example, Donald Trump is likely to elicit a negative moral judgement from most Liberals and a positive one from most Conservatives, regardless of the specific moral foundation that is evoked. From a technical perspective, our goal is to model these kind of dependencies in a probabilistic framework, connecting MF and roles assignments, entity-specific sentiment polarity and repeating patterns within ideological groups (while our focus is U.S. politics, these settings could be easily adapted to capture patterns based on other criteria). We formulate these dependencies as a structured learning task and compare two relational learning frameworks, PSL~\cite{bach:jmlr17} and DRaiL~\cite{pacheco-goldwasser-2021-modeling}. Our experiments demonstrate that modeling these dependencies, capturing political and social knowledge, result in improved performance. In addition, we conduct a thorough ablation study and error analysis to explain their impact on performance. 


Finally, we demonstrate how entity-based MF analysis can help capture perspective differences based on ideological lines. We apply our model to tweets by members of Congress on the issue of Abortion and the 2021 storming of the US Capitol. Our analysis shows that while Conservative and Liberal tweets target the same entities, their attitudes are often conflicting.

%% file: 2related.tex
\section{Related Work}

Usage of sociological theories like the Moral Foundation Theory (MFT)~\citep{haidt2004intuitive,haidt2007morality} and Framing~\citep{entman1993framing,chong2007framing,boydstun2014tracking} in Natural Language Processing tasks has gained significant interest. The Moral Foundation Theory (MFT) has been widely used to study the effect of moral values on human behavioral patterns, such as charitable donations \cite{hoover2018moral}, violent protests \cite{mooijman2018moralization} and social homophily \cite{dehghani2016purity}. Framing is a strategy used to bias the discussion on an issue towards a specific stance by emphasizing certain aspects that prime the reader to support the stance. Framing is used to study the political bias and polarization in social and news media~\citep{tsur-etal-2015-frame,baumer-etal-2015-testing,card-etal-2015-media,field-etal-2018-framing,demszky2019analyzing,fan-etal-2019-plain,roy-goldwasser-2020-weakly}. Moral Foundation Theory (MFT) is frequently used to analyze political framing and agenda setting. For example, \citet{fulgoni-etal-2016-empirical} analyzed framing in partisan news sources using MFT, \citet{dehghani2014analyzing} studied the difference in moral sentiment usage between liberals and conservatives. \citet{brady2017emotion} found that moral/emotional political messages are diffused at higher rates on social media. 

Previous works have also contributed to the detection of moral sentiments. \citet{johnson-goldwasser-2018-classification} showed that policy frames \cite{boydstun2014tracking} help in moral foundation prediction, \citet{hoover2020moral} proposed a dataset of $35$k tweets annotated for moral foundations, \citet{lin2018acquiring} used background knowledge for moral sentiment prediction, \citet{xie2020text} proposed a text based framework to account for moral sentiment change, and \citet{garten2016morality} used pretrained distributed representations of words to extend the Moral Foundations Dictionary~\cite{DVN/SJTRBI_2009} for detecting moral rhetoric. 

While existing works study MFT at the issue and sentence level, \citet{roy-goldwasser-2021-analysis} showed that there is a correlation between entity mention and the sentence-level moral foundation in the tweets by the U.S. politicians. We extend this work by studying MFT directly at the entity level. Hence, our work is broadly related to the works on entity-centric affect analysis~\citep{deng-wiebe-2015-joint,field2019entity,park2020multilingual}.

Combining neural networks and structured inference was explored for traditional NLP tasks such as dependency parsing~\cite{chen-manning-2014-fast,weiss-etal-2015-structured,andor-etal-2016-globally}, named entity recognition~\cite{lample-etal-2016-neural} and sequence labeling systems~\cite{ma-hovy-2016-end,zhang-etal-2017-semi}. Recently, these efforts have expanded to discourse-level tasks such as argumentation mining~\cite{niculae-etal-2017-argument,widmoser-etal-2021-randomized}, event/temporal relation extraction~\cite{han-etal-2019-joint} and discourse representation parsing~\cite{liu-etal-2019-discourse}. Following this trend, \citet{pacheco-goldwasser-2021-modeling} introduced DRaiL, a general declarative framework for deep structured prediction, designed specifically for NLP tasks. In this paper, we use DRaiL to model moral foundations and morally-targeted entities in tweets, and find an improvement over other non-neural probabilistic graphical modeling frameworks \cite{bach:jmlr17}.

%% file: 3data.tex
\vspace{-3mm}
\begin{table}[ht!]
\begin{center}
 \scalebox{0.70}{\begin{tabular}{>{\arraybackslash}m{5.5cm}|>{\arraybackslash}m{4.5cm}} 
 \hline 
 
\textsc{\textbf{Moral Foundations}} & \textsc{\textbf{Moral Roles}}\\ [0.5ex]
\hline
\textsc{\textbf{Care/Harm:}} Care for others, generosity, compassion, ability to feel pain of others, sensitivity to suffering of others, prohibiting actions that harm others. & 
\makecell[l]{1. Target of care/harm\\2. Entity causing harm\\3. Entity providing care}\\
\hline
\textsc{\textbf{Fairness/Cheating:}} Fairness, justice, reciprocity, reciprocal altruism, rights, autonomy, equality, proportionality, prohibiting cheating. & 
\makecell[l]{1. Target of fairness/cheating\\2. Entity ensuring fairness\\3. Entity doing cheating}\\
\hline
\textsc{\textbf{Loyalty/Betrayal:}} Group affiliation and solidarity, virtues of patriotism, self-sacrifice for the group, prohibiting betrayal of one’s group. & 
\makecell[l]{1. Target of loyalty/betrayal\\2. Entity being loyal\\3. Entity doing betrayal}\\
\hline
\textsc{\textbf{Authority/Subversion:}} Fulfilling social roles, submitting to authority, respect for social hierarchy/traditions, leadership, prohibiting rebellion against authority. & 
\makecell[l]{1. Justified authority\\2. Justified authority over\\3. Failing authority\\4. Failing authority over}\\
\hline
\textsc{\textbf{Purity/Degradation:}} Associations with the sacred and holy, disgust, contamination, religious notions which guide how to live, prohibiting violating the sacred. & 
\makecell[l]{1. Target of purity/degradation\\2. Entity preserving purity\\3. Entity causing degradation}\\
\hline
\end{tabular}}
\caption{Moral foundations and their associated roles.}
\label{tab:moral-foundation-roles}
\end{center}
\end{table}

\section{Identifying Entity-Centric Moral Roles}\label{sec:data}

\subsection{Morality Frames}\label{sec:identification_entity_role} MFT defines a convenient abstraction of the moral sentiment expressed in a given text. \textbf{Morality Frames} build on MFT and provide \textit{entity-centric moral sentiment} information. Rather than defining negative and positive MF polarities (e.g., CARE or HARM), we use the five MFs as frame predicates, and associate positive and negative entity roles with each frame. As described in Table~\ref{tab:moral-foundation-roles}, these roles capture information specific to each MF. For example, \textit{entity causing harm}, is a negative sentiment role, associated with the CARE/HARM morality frame. The entities filling these roles can be individuals, collective entities, objects, activities, concepts, or legislative elements. 


\subsection{Data Collection}\label{sec:data_collection}
We build on the dataset proposed by \citet{johnson-goldwasser-2018-classification}, 
 consisting of tweets by US politicians posted between 2016 and 2017. A subset of it (2K out of 93K) is annotated for Moral Foundations and Policy Frames \cite{boydstun2014tracking}. The tweets focus on six politically polarized issues: \textit{immigration, guns, abortion, ACA, LGBTQ, and terrorism}, and the party affiliations of the authors are known. 
We consider only labeled moral tweets, and choose the most prominent MF annotation for each tweet (some tweets are annotated for a secondary MF). 
Since the data contains only few examples of the \textit{Purity/Degradation} moral foundation, 
we collected more examples from the unlabeled segment and manually annotated them. Table \ref{tab:dataset_summary} shows the statistics of the final dataset.
The annotation process and per-topic distribution of tweets are outlined in Appendix \ref{appendix:dataset}. 

\begin{table}[ht!]
\begin{center}
 \scalebox{0.7}{\begin{tabular}{>{\arraybackslash}m{3.5cm}|>{\centering\arraybackslash}m{1.3cm}||>{\centering\arraybackslash}m{1cm}|>{\centering\arraybackslash}m{1cm}} 
 \hline 
\textsc{\textbf{Morals}} & \textsc{\textbf{\# of Tweets}} & \multicolumn{2}{c}{\textsc{\textbf{Ideology}}} \\ [0.5ex]
\cline{3-4}
&   & \textsc{Left}   &   \textsc{Right}  \\
 \hline
 Care/Harm              & 589 & 378 & 211 \\
 Fairness/Cheating      & 264 & 201 & 63  \\
 Loyalty/Betrayal       & 231 & 167 & 64  \\
 Authority/Subversion   & 471 & 200 & 271 \\
 Purity/Degradation     & 44  & 13  & 31  \\
 \hline 
 \textsc{\textbf{Total}}& 1599& 959 & 640 \\
 \hline
 \end{tabular}}
\caption{Dataset summary.}
\label{tab:dataset_summary}
\end{center}
\end{table}

\subsection{Entity Roles Annotation}\label{sec:annotation_entity_roles}
We annotate each tweet for entities and their  associated moral roles. 

\vspace{5pt}
\noindent\textbf{Annotation Schema:} We set up a QA task on Amazon Mechanical Turk. Annotators were given a tweet, the associated MF label and its description. They were then presented with multiple questions, and asked to mark the answers, corresponding to our entity roles, in the tweet. Table~\ref{tab:questionnaire-mturk-care} shows the questions asked for the \textit{Care/Harm} case. We asked additional questions to assess the annotators' understanding of the task. The questions for other moral foundations can be found in Appendix \ref{appendix:questionnaire-mturk-full}. 
\begin{table}[t!]
\begin{center}
\scalebox{0.7}{
\begin{tabular}{>{\arraybackslash}m{3.1cm}|>{\arraybackslash}m{7cm}}
    \hline
    \textbf{\textbf{Entity Type}} & \textbf{Question Asked to the Annotators}\\ [0.5ex]
    \hline

    Target of care/harm 
    & Which entity needs care, or is being harmed?\\
    \hline
    Entity causing harm 
    & Which entity is causing the harm?\\
    \hline
    Entity provid. care 
    & Which entity is offering/providing the care?\\
    \hline

\end{tabular}
}
\caption{Questionnaire for entity roles in care/harm.}
\label{tab:questionnaire-mturk-care}
\vspace{0.2cm}
 \scalebox{0.64}{\begin{tabular}{>{\arraybackslash}m{3.2cm}||>{\centering\arraybackslash}m{0.6cm}|>{\centering\arraybackslash}m{0.6cm}||>{\centering\arraybackslash}m{0.6cm}|>{\centering\arraybackslash}m{0.6cm}||>{\centering\arraybackslash}m{1.5cm}|>{\centering\arraybackslash}m{1.5cm}} 
 \hline 
\textbf{\textsc{Morals}} & \multicolumn{2}{c}{\textsc{\textbf{\# Tweets}}} & \multicolumn{2}{c}{\textsc{\textbf{\# Ann/Tw}}} & \multicolumn{2}{c}{\textsc{\textbf{Agreement (SD)}}}\\ [0.5ex]
\cline{2-7}
&   \textsc{S}   &   \textsc{F}   &   \textsc{S}   &   \textsc{F}   &   \textsc{S}  &   \textsc{F}\\
 \hline
 Care/Harm             & 27  & 589  & 3   & 3   &   0.63 (0.5)  &   0.70 (0.5) \\
 Fairness/Cheating     & 30  & 247  & 5.03  & 2.92    &   0.55 (0.4)  &   0.69 (0.5) \\
 Loyalty/Betrayal      & 40  & 203  & 5.67   & 2.89   &   0.58 (0.3)  &   0.63 (0.5) \\
 Authority/Subversion  & 50  & 466  & 4.58  & 2.92   &   0.55 (0.3)  &   0.60 (0.5) \\
 Purity/Degradation    & 10  & 36   & 6     & 3   &   0.51 (0.4)  &   0.77 (0.6) \\
 \hline
 \end{tabular}}
\caption{Annotator agreement in selection (S) and final (F) phases. \textsc{Ann/Tw} refers to number of annotations per tweet and SD refers to Standard Deviation.
}
\label{tab:annotation-agreement}
\end{center}
\vspace{-5mm}
\end{table}

\noindent\textbf{Quality Assurance:} 
We provided the annotators with work-through examples and hints with each question about the entity type. The interface allowed them to mark a segment of the text with one moral role only. To further improve the quality, we did the annotation in two phases. In the annotator selection phase, we released a small subset of tweets for annotation. Based on the annotations, we assigned qualifications to high performing workers and released the rest of the tweets only to them. We awarded the annotators $15-18\cent$ per tweet. We define agreement among annotators if they mark the same segment in the text as having the same entity-role. We calculate the agreement among multiple annotators using Krippendorff’s $\alpha$ \cite{krippendorff2004measuring}, where $\alpha=1$ means perfect agreement, $\alpha=-1$ means inverse agreement, and $\alpha=0$ is the level of chance in a tweet. Table \ref{tab:annotation-agreement} shows that the average agreement increased in the final stage. Note that the annotator agreement (Krippendorff’s $\alpha$) is calculated by comparing the character by character agreement between annotations. For example, if one annotator has marked `President Trump' as an answer in a tweet, and another has marked `Trump' as the answer, it will be considered as agreement on the characters `Trump' but disagreement on `President', although they really did not disagree on their annotations. This makes the agreement measurement very strict. Regardless, we still got very good average agreement among annotators in the final annotation step. We further refine the annotations by taking majority voting as described in the following section.

\noindent\textbf{Annotation Results:} A tweet is annotated by at least three annotators. We define a text span to be an entity E, having a moral role M, in tweet T, if it is annotated as such by at least two annotators. This way, we found $2,945$ (T, E, M) tuples. 

To compare the partisanship of MFs and MF roles, we calculate the z-scores for the proportion of MFs and MF roles in the left and right, and consider it as partisan score (- right, + left). The partisan scores for common MFs and their corresponding most partisan (role: entity) tuples are shown in Table \ref{tab:partisanship}. The results of this analysis align with our intuition, moral sentiment towards entities can be more indicative of partisanship than the high-level MFs. 
In Table \ref{tab:example-entities-care}, we present the top-5 most used entities per role by political party for \textit{Care/Harm}. We can see that the target entities of moral roles vary significantly across parties. Details for other MFs and z-scores are in Appx. \ref{appendix:most-frequent-entities-by-entity-role}.

\begin{table}[t]
\begin{center}
\scalebox{0.65}{\begin{tabular}{>{\arraybackslash}m{1cm}|>{\raggedright}m{2.7cm}||>{\arraybackslash}m{3.1cm}|>{\arraybackslash}m{3cm} } 
 \hline 
\multirow{2}{*}{\textbf{Topic}} & \textbf{Common MF} & \multicolumn{2}{c}{\textbf{Most Partisan Role:Entity}} 
\\ 
~ & \textbf{(Partisan score)} & \textbf{Right (-)} & \textbf{Left (+)} \\
 \hline
 Abort & Care/Harm (+1.4) & \makecell[l]{Harming:\\PPFA (-3.2)} & \makecell[l]{Caring:\\PPFA (+0.5)} \\
 \hline
 Immig & Auth/Subv (-6.8) & \makecell[l]{Failing Authority:\\Obama (-0.5)} & \makecell[l]{Failing Authority:\\SCOTUS (+5.3)}  \\
 \hline
 Guns & Care/Harm (+0.6) & \makecell[l]{Care For: Law Abid-\\ing Citizens (-4.8)} & \makecell[l]{Harming:\\Gun (+4.1)}\\
 \hline 
 ACA & Care/Harm (+2.2) & \makecell[l]{Harming:\\ACA (-5.7)} & \makecell[l]{Caring:\\ACA (+5.6)}\\
 \hline 
 Terror & Care/Harm (+2.0) & \makecell[l]{Harming:\\Islam (-2.4)} & \makecell[l]{Harming:\\Terr. Suspect (+3.5)}\\
 \hline
 LGBT & Fair/Cheat (+2.1) & \makecell[l]{Cheating: SCOTUS-\\Marriage (-6.5)} & \makecell[l]{Target of Fairness:\\LGBT (+1.0)}\\
 \hline
 \end{tabular}}
\caption{Partisanship of MF and MF Roles. 
}
\label{tab:partisanship}
\end{center}
\vspace{-5mm}
\end{table}

\begin{table}[t]
\begin{center}
\scalebox{0.7}{
\begin{tabular}{>{\raggedright\arraybackslash}m{1.5cm}|>{\arraybackslash}m{4cm}|>{\arraybackslash}m{4cm}}
    \hline
    \textbf{Entity Types} & \textbf{Most Frequent Entities in Left} & \textbf{Most Frequent Entities in Right}\\ [0.5ex]
    \hline
    Target of care/harm 
    & 20 million Americans; our families; woman; innocent people; \#domesticviolence victims 
    & law-abiding Americans; victims and their families; small businesses; patients; Paris\\
    \cline{1-3}
    Entity causing harm 
    & gun show loopholes; gun violence; terrorist attack; mass-shootings; suspected terrorists 
    & Radical Islamic terrorists; \#Obamacare mandates; Brussels attacks; \#ISIS; ISIL-Inspired Attacks\\
    \cline{1-3}
    Entity providing care 
    & gun show loophole bills; Affordable Care Act; \#ImmigrationReform; Democrats; commonsense gun legislation 
    & @RepHalRogers: Bill; @HouseGOP; Senate; @WaysandMeansGOP; HR 240\\
    \hline
\end{tabular}
}
\caption{Top-5 frequent entities by role for  {Care/Harm}.}
\label{tab:example-entities-care}
\end{center}
\end{table}

    


%% file: 4model.tex
\section{Model}\label{sec:model}
We propose a relational learning model for identifying morality frames in text. We begin by defining our relational structure  (Sec.~\ref{sec:rules-constraints}) and proceed to describe its implementation using relational learning tools (Sec.~\ref{sec:relational-framework}).
\subsection{Relational Model for Morality Frames}\label{sec:rules-constraints} 
Statistical Relational Learning (SRL) methods attempt to model a joint distribution over relational data, and have proven useful in tasks where contextualizing information and interdependent decisions can compensate for  a low number of annotated examples~\citep{deng-wiebe-2015-joint,johnson2016identifying,johnson-goldwasser-2018-classification,subramanian2018hierarchical}. 
By breaking down the problem into interdependent relations, these approaches are easier to interpret than end-to-end deep learning techniques. 

We propose a joint prediction framework of morality frames, modeling the dependency between MF labels and moral roles instances. Following SRL conventions~\cite{richardson2006markov,bach:jmlr17}, we use first-order-logic to describe relational properties. Specifically, a logical rule is used to define a probabilistic scoring function over the relation instances appearing in it, the full description appears in Section \ref{sec:relational-framework}.
\vspace{-2mm}
\begin{equation*}\small
\begin{split}
& r_1: \mathtt{Tweet(t)} \Rightarrow \mathtt{MF(t,m)} \\
& r_2: \mathtt{Tweet(t)} \wedge \mathtt{Ent(t,e)} \Rightarrow 
\mathtt{Role(t,e,r)} \\
\end{split}
\end{equation*}
%
In addition, we make the observation that both moral foundations and entities' moral roles depend on external factors that go beyond the text, such as author information and party affiliation. Previous work has shown that explicitly modeling party affiliation and the topics discussed are helpful for predicting moral foundations \cite{johnson-goldwasser-2018-classification}. For this reason, we condition both the moral foundation and moral roles on this additional information, as shown in the rules below.
\begin{equation*}\small
\begin{split}
& r_3: \mathtt{Tweet(t)} \wedge \mathtt{Ideo(t,i)} \wedge \mathtt{Topic(t,k)} \Rightarrow \mathtt{MF(t,m)} \\
& r_4: \mathtt{Tweet(t)} \wedge \mathtt{Ideo(t,i)} \wedge \mathtt{Topic(t,k)} \wedge \mathtt{Ent(t,e)} \Rightarrow \mathtt{Role(t,e,r)}  \\
\end{split}
\end{equation*}

Rules $r_1$, $r_2$ condition the moral foundation label ($\mathtt{m}$) and moral foundation role label ($\mathtt{r}$) on the tweet ($\mathtt{t}$) and entity ($\mathtt{e}$), while $r_3$, $r_4$ condition on the ideology of the author ($\mathtt{i}$) and the topic of the tweets ($\mathtt{k}$). Concretely, $r_4$ can be translated as \textit{``if a tweet $\mathtt{t}$ has author ideology $\mathtt{i}$, topic $\mathtt{k}$, and mentions entity $\mathtt{e}$, the entity will have moral role $\mathtt{r}$''}. Other rules can be translated similarly. Then, we explicitly model the dependencies among different decisions using the following three constraints.
\begin{equation*}\small
\begin{split}
& c_1: \mathtt{Ent(t,e)} \wedge \mathtt{Role(e,r)} \wedge \mathtt{MF\_Role(m,r)} \Rightarrow \mathtt{MF(t,m)} \\
& c_2: \mathtt{Ent(t,e_1)} \wedge \mathtt{Ent(t,e_2)} \wedge \mathtt{Role(t, e_1,r)}  \Rightarrow \mathtt{\neg Role(t,e_2,r)} \\
& c_3: \mathtt{SameIdeo(t_1,t_2)} \wedge \mathtt{SameTopic(t_1,t_2)} \wedge \mathtt{Ent(t_1,e)} \wedge \mathtt{Ent(t_2,e)} \\
& \;\;\;\;\;   \wedge \mathtt{Role(t_1,e,r_1)} \wedge \mathtt{Role(t_2,e,r_2)} \Rightarrow \mathtt{SamePolarity(r_1, r_2)} \\
\end{split}
\end{equation*}

\textbf{($c_1$) Consistency between MF label and roles:} While rules $r_1$, $r_3$ predict the MF labels, and $r_2$, $r_4$ predict the role labels, these two predictions are interdependent. Knowing the MF of a tweet limits the space of feasible roles. Likewise, knowing the role of an entity in a tweet will directly give us the MF label. For example, the presence of an entity frequently used as a harming entity indicates a higher probability of the MF label `Care/Harm'. We model the dependency between these two decisions using constraint $c_1$, which can be translated as \textit{``if an entity $\mathtt{e}$, mentioned in tweet $\mathtt{t}$, has role $\mathtt{r}$, tied to MF $\mathtt{m}$, then tweet $\mathtt{t}$ will have MF label $\mathtt{m}$''. }

\textbf{($c_2$) Different roles for different entities in the same tweet:} Our intuition is that if multiple entities are mentioned in the same tweet, they are likely to have different roles. While this may not always hold true, we use this constraint to prevent the model from relying only on textual context, and assigning the same role to all entities.

\textbf{($c_3$) Consistency in the polarity of sentiment towards an entity within a political party:} Intuitively, role types have a positive or negative sentiment associated to them. For example, an entity causing harm and an entity doing betrayal carry negative sentiment. Intuitive polarity for each MF role can be found in Appendix \ref{appendix:moral-role-polarity}. Given the highly polarized domain that we are dealing with, we assume that regardless of the MF, an entity will likely maintain the same polarity when mentioned by a specific political party across the same topic. Constraint $c_3$ encourages this consistency, and it can be translated as: \textit{``if two tweets $\mathtt{t_1}$, $\mathtt{t_2}$ are written by authors of the same political ideology, on the same topic, and mention the same entity $\mathtt{e}$, then the polarity of the roles $\mathtt{r_1}$ and $\mathtt{r_2}$ of $\mathtt{e}$ in both tweets will likely be the same.''}   We consider two entities to be the same if they are an exact lexical match, and leave entity clustering for future work. 

\subsection{Frameworks for Relational Learning}\label{sec:relational-framework} In this work, we experiment with two existing frameworks for modeling relational learning problems - (1) Probabilistic Soft Logic~(PSL) \cite{bach:jmlr17} and (2) Deep Relational Learning (DRaiL)~\cite{pacheco-goldwasser-2021-modeling}. Both PSL and DRaiL are probabilistic frameworks for specifying and learning relational models using weighted logical rules, specifically horn clauses of the form $w_r: P_1 \wedge ... \wedge P_{n-1} \Rightarrow P_n$. Weights $w_r$ indicate the importance of each rule in the model, and they can be learned from data. Predicates $P_i$ can be closed, if they are observed, or open if they are unobserved. Probabilistic inference is used over all rules to find the most probable assignment to open predicates. The main differences between PSL and DRaiL are - (a) In DRaiL, each rule weight is learned using a neural network, which can take arbitrary input representations, while in PSL a single weight is learned for each rule, and expressive classifiers can only be leveraged as priors; (b) DRaiL defines a shared relational embedding space, by specifying entity and relation specific encoders that are reused across all rules. In both frameworks, rules are transformed into linear inequalities corresponding to their disjunctive form, and MAP inference is defined as a linear program. 

In PSL, rules are compiled into a Hinge-Loss Markov random field, defined over continuous variables. Weights can be learned using maximum likelihood estimation, maximum-pseudolikelihood estimation, or large-margin estimation. In DRaiL,
rule weights are learned using neural networks. Parameters can be learned \textit{locally}, by training each neural network independently, or \textit{globally}, by using inference to ensure that the scoring functions for all rules result in a globally consistent decision. To learn global models, DRaiL can also employ maximum likelihood estimation or large-margin estimation. Details regarding both frameworks can be found in Appendix \ref{app:reln_models}.


%
%

%% file: 5eval.tex
\section{Experimental Evaluation}\label{sec:eval}

The goal of our relational learning framework is to identify morality frames in tweets by modeling them jointly, and derive interpretable relations between them and other contextualizing information. In this section, we compare the performance of our model with multiple baselines, and present a detailed error analysis. Then, we collect tweets on one topic (\textit{Abortion}) and one event (\textit{2021 US Capitol Storming}) written by US Congress members and analyze the discussion.\footnote{Collected from https://github.com/alexlitel/congresstweets} We identify the morality frames in these tweets using our best model. 

\subsection{Experimental Settings}
We experiment with PSL and DRaiL for modeling the rules presented in Section \ref{sec:rules-constraints}. In DRaiL, each rule $r$ is associated with a neural architecture, which serves as a scoring function to obtain the rule weight $w_r$. In the case of rules $r_1$ and $r_2$, which map tweets and entities to labels, we use a BERT encoder \cite{devlin-etal-2019-bert} with a classifier on top. 
We use task-adaptive pretraining for BERT~\cite{gururangan-etal-2020-dont}, and fine-tune it on a large number of unlabeled tweets
. In the case of rules $r_3$ and $r_4$, that incorporate ideology and topic information, we learn topic and ideology embeddings with one-layer feed-forward nets over their one-hot representations. Then, we concatenate the output of BERT with the topic and ideology embeddings before passing everything through a classifier. On the other hand, PSL directly learns a single weight for each rule. Given that our rules are defined over complex inputs (tweets), we use the output of the locally trained neural nets as priors for PSL, by introducing additional rules of the form $\mathtt{Prior(t,x)} \Rightarrow \mathtt{Label(t,x)}$. This approach has been successfully used in previous work dealing with textual inputs \cite{sridhar-etal-2015-joint}. Note that while PSL can only leverage these classifiers as priors, DRaiL continues to update the parameters of the neural nets during learning. 

We model constraint $c_1$, aligning the MF and role predictions, and $c_3$, aligning role polarity, as unweighted hard constraints in both frameworks. 
For constraint $c_2$, we learn a weight to encourage different entities in a tweet to have different roles. PSL learns a weight directly over this rule, while in DRaiL we use a feed-forward net over the one-hot vector of the relevant MF. We compare our relational models with the following baselines.

\textbf{Lexicon Matching:} Direct keyword matching using the MF Dictionary (MFD)~\cite{DVN/SJTRBI_2009} and a PMI-based lexicon extracted from the dataset by \citet{johnson-goldwasser-2018-classification}.

\textbf{Sequence Tagging:} We set the MF role prediction task as a sequence tagging problem, and map each entity in a tweet to a role label. We use a BiLSTM-CRF~\cite{huang2015bidirectional} over the full tweet, and use the last time-step in each entity span as its emission probability. 

\textbf{End-to-end Classifiers:} We map the text and entities, and other contextualizing features (e.g. topic), to a single label. We compare BERT-base and task adaptive pretraining (BERT-tapt) by using a whole-word-masking objective over the large set of unlabeled political tweets. 

\textbf{Multi-task}: We define a single BERT encoder, and a single ideology and topic embedding that is shared across the two tasks. Task-specific classifiers are used on top of these representations. Then, the loss functions are added as $L = \lambda_1 L_{\mathtt{MF}} + \lambda_2 L_{\mathtt{Role}}$. We set $\lambda_1 = \lambda_2 = 1$.

\begin{table}[t!]
    \centering
    \resizebox{\columnwidth}{!}{%
    \begin{tabular}{llcccc}
    \toprule
    \multirow{2}{*}{\textbf{\textsc{Group}}} & \multirow{2}{*}{\textbf{\textsc{Model}}} & \multicolumn{2}{c}{\textbf{\textsc{Macro}}} & \multicolumn{2}{c}{\textbf{\textsc{Weighted}}} \\
        ~ & ~ & \textsc{Role} & \textsc{MF} & \textsc{Role} & \textsc{MF} \\
        \midrule
    Lexicon & MF Dictionary & - & 30.37 & - & 37.32 \\
    Matching & PMI Lexicon & - & 36.44 & - & 35.94 \\
    ~ & MFD + PMI & - & 39.78 & - & 42.12 \\
    \midrule
    \multirow{1}{*}{Seq-Tagging} & BiLSTM-CRF & 35.18 & - & 45.91 & - \\
    \midrule
                     & BiLSTM & 39.75 & 58.61 & 45.61 & 59.90  \\
    End-to-end       & BERT-base  & 49.32 & 59.99 & 57.37 & 62.17 \\
    Classifiers       & BERT-tapt  & 54.73 & 66.44 & 62.18 & 68.29 \\
           & + Ideo + Issue &  54.81 & 66.13 & 62.83 & 68.34  \\
    \midrule
    \multirow{3}{*}{Multi-task} & BERT-base & 44.37 & 61.63 & 57.74 & 67.71 \\
    ~ & BERT-tapt & 52.08 & 63.46 & 61.96 & 69.20 \\
    & + Ideo + Issue & 52.11 & 63.44 & 63.61 & 68.61 \\
    \midrule
    Relational & PSL & 56.51 &  68.98 & 64.02 & 71.85 \\
    Learning & DRaiL Local   & 58.07 & 71.20 & 64.38 & 73.85 \\
    & DRaiL Global     & \textbf{59.23  }  & \textbf{72.34} & \textbf{64.98} & \textbf{74.39}  \\
    \midrule
    \multirow{2}{*}{Skyline} & DRaiL Global & \multirow{2}{*}{79.35} & \multirow{2}{*}{-} & \multirow{2}{*}{84.52} & \multirow{2}{*}{-} \\
    
    & (Fixed MF) \\
    \bottomrule
    \end{tabular}
    }
    \caption{MF and MF role classification F1 Scores.}
    \label{tab:general}
    \vspace{0.2cm}

    \centering
    \resizebox{1\columnwidth}{!}{%
    \begin{tabular}{lccccccc}
    \toprule
      \multirow{3}{*}{\textbf{\textsc{Models}}} &
      \multicolumn{2}{c}{\textbf{\textsc{Weighted F1}}} & \multicolumn{3}{c}{\textbf{\textsc{\# of Errors}}} \\
      \cmidrule(lr){2-3}\cmidrule(lr){4-6}
        ~ & \multirow{2}{*}{Role} & \multirow{2}{*}{MF} & Polarity & Mixed & Same \\
        ~ &    &   & Swap (E1) & MFs (E2) & Role (E3) \\
        \midrule
        BERT-tapt                           &  62.18 & 68.29    & 274 & 844 & 102\\
        \midrule
        All rules    & 63.93 & 69.23    & 260 & 807 & 93\\
        $+c_1$             &  65.11 & 74.44    & 254 & \textbf{732} & 130\\
        $+c_2$             & 63.94 & 69.37    & 260 & 790 & 101\\
        $+c_3$             & 64.13 & 69.31    & 245 & 797 & \textbf{89}\\
        $+c_1+c_2$         & 65.04 & 74.43    & 254 & 733 & 149\\
        $+c_1+c_3$         & \textbf{65.32} & \textbf{74.53} & 249 & 733 & 126\\
        $+c_2+c_3$         & 63.99 & 69.22 & \textbf{244} & 791 & 96 \\
        + All constr    & 64.98 & 74.39 & 248 & 736 & 138  \\
        \bottomrule
    \end{tabular}}
    \caption{Ablation Study and Error Analysis. 
    }\label{tab:ablation-error}
\vspace{-4mm}
\end{table}

We perform 3-fold cross validation over the dataset introduced in Section \ref{sec:data}, and show results for MF and role prediction in Table \ref{tab:general}. First, we observe that leveraging unlabeled data for task-adaptive pretraining improves performance. Then, we find that relational models that use probabilistic inference outperform all of the other baselines for both tasks. Further, we find that modeling rules using neural nets in DRaiL, and learning their parameters with global learning, performs better than using them as priors and learning a single weight in PSL. We also include results by fixing the gold labels for the MF prediction, and refer to this as a \textit{skyline}. Unsurprisingly, having perfect MF information improves results for roles considerably. In this case, the candidates for each entity are reduced from 16 possible assignments to 3 or 4, which results in a much easier task. Details regarding all baselines, hyper-parameters, task-adaptive pretraining, and results per class can be found in Appendix \ref{appendix:experimental-eval}. Code and dataset can be found at \url{https://github.com/ShamikRoy/Moral-Role-Prediction}.

\subsection{Ablation Study and Error Analysis} We perform an ablation study, evaluating different versions of our model by adding and removing constraints and analyzing corresponding errors. To study the effect of different rules and constraints on role prediction, we define three types of errors:

\textbf{(E1) Polarity Swap}: when the role of an entity with one polarity (positive/negative) is identified as one role of the opposite polarity.

\textbf{(E2) Mixed MFs}: when different entities of the same tweet are identified with roles from a MF other than the gold label of the tweet.

\textbf{(E3) Same Roles}: all of the entities in a tweet are identified to have the same role when the gold labels are different. 

The analysis is shown in Table \ref{tab:ablation-error}. First, we see that constraint $c_1$, aligning the two decisions, does most of the heavy lifting and reduces error (E2) in all cases. 
Enforcing consistent polarities with $c_3$ further improves performance and reduces error (E1), for which it is designed for. $c_3$ also reduces error (E3) in some models. Encouraging entities to have different roles with $c_2$ does not improve the overall performance, but it helps to reduce error (E3) when combined with $c_3$. We use a soft version of $c_2$, so it is not strictly enforced. We find that roles with negative sentiments are easier for the model to identify (Appendix \ref{app:per_class}). Note that every MF has only one role with negative sentiment, and the model does not swap role labels with different sentiments frequently (E1). Therefore, determining the correct positive role is more challenging. 

\subsection{Predicting Morality Frames on New Data}
To analyze the political discussion using the moral sentiment towards entities, we collected more tweets from US politicians on the topic of Abortion and around the storming of the US Capitol on Jan. 6, 2021. The Abortion tweets are from 2017 to Feb. 2021. For the US Capitol incident, we collected tweets 7 days before and after the event, with the goal of studying any change in sentiment towards entities. 
We took noun phrases occurring at least $50$ times, manually filtered out non-entities, and grouped different mentions of the same entity (Appendix \ref{appendix:entity-groups}). We collected tweets that mentioned these entities. Statistics for the resulting data can be found in Table \ref{tab:new_set}. We re-trained our model using all of our labeled data, and predicted the morality frames for each tweet in the new dataset.


\begin{table}
\centering
    \resizebox{0.9\columnwidth}{!}{%
\begin{tabular}{llllll}
\toprule
    \multirow{2}{*}{Topic} & \multirow{2}{*}{Tweets} & \multirow{2}{*}{NPs}  & \multirow{2}{*}{Ents} & Final  &  Final  \\
    ~     &        &      &    & Tweets & (Tw, Ent) \\
    \midrule
    \textbf{\textsc{Abortion}}   & 18.5k & 156 & 25 & 9,676 & 28,054 \\
    \textbf{\textsc{US Capitol}} & 22.2k & 100 & 25 & 7,188 & 14,299 \\
    \bottomrule
\end{tabular}}
\caption{Summary of new dataset from US politicians.}\label{tab:new_set}
\vspace{6pt}
\resizebox{0.9\columnwidth}{!}{%
\begin{tabular}{llllll}
\toprule
    \multicolumn{3}{c}{\textbf{\textsc{Abortion}}} &  \multicolumn{3}{c}{\textbf{\textsc{US Capitol}}} \\
    \cmidrule(lr){1-3}\cmidrule(lr){4-6}
    Entity & Freq. & Prec. & Entity & Freq. & Prec. \\
   \cmidrule(lr){1-3}\cmidrule(lr){4-6}
    Women & 91 & 0.89 & Trump & 2 & 0.50 \\
    P. Parenthood & 60 & 0.75 & Congress & 58 & 0.58\\
    Life & 31 & 0.50 & Capitol & 1 & 0.57\\
    \bottomrule
\end{tabular}}
\caption{Precision of moral role prediction of entities in new data vs. entity frequency in training data.}\label{tab:train_freq}
\end{table}

\begin{figure*}[t!]
    \begin{center}
    \begin{subfigure}[t]{0.3\textwidth}
        \centering
        \includegraphics[width=1\textwidth]{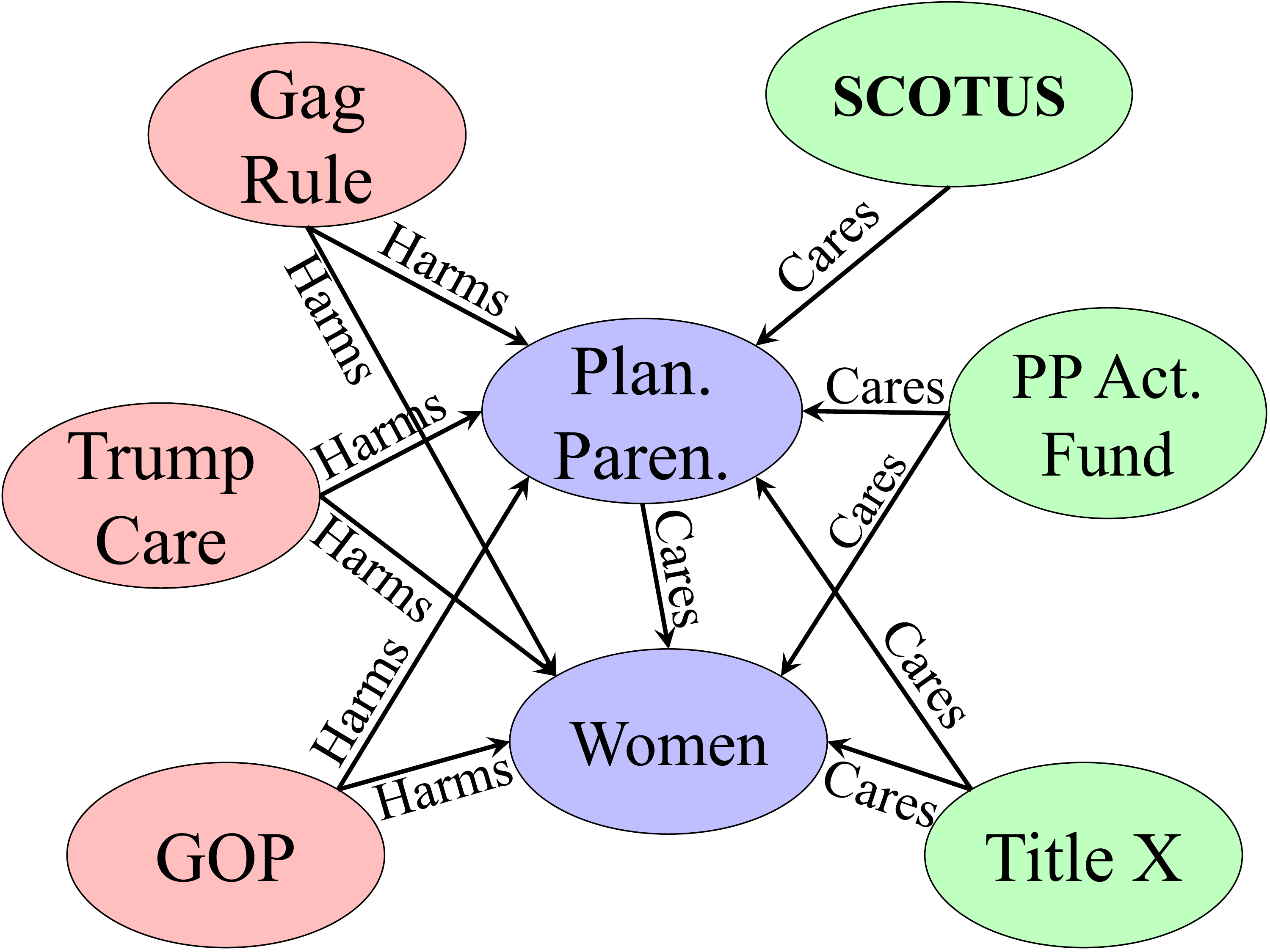}
        \caption{Entity relationship in left}
        \label{fig:entity-relation-left-abortion}
    \end{subfigure}
    \hspace{10mm}%
    \begin{subfigure}[t]{0.3\textwidth}
        \centering
        \includegraphics[width=1\textwidth]{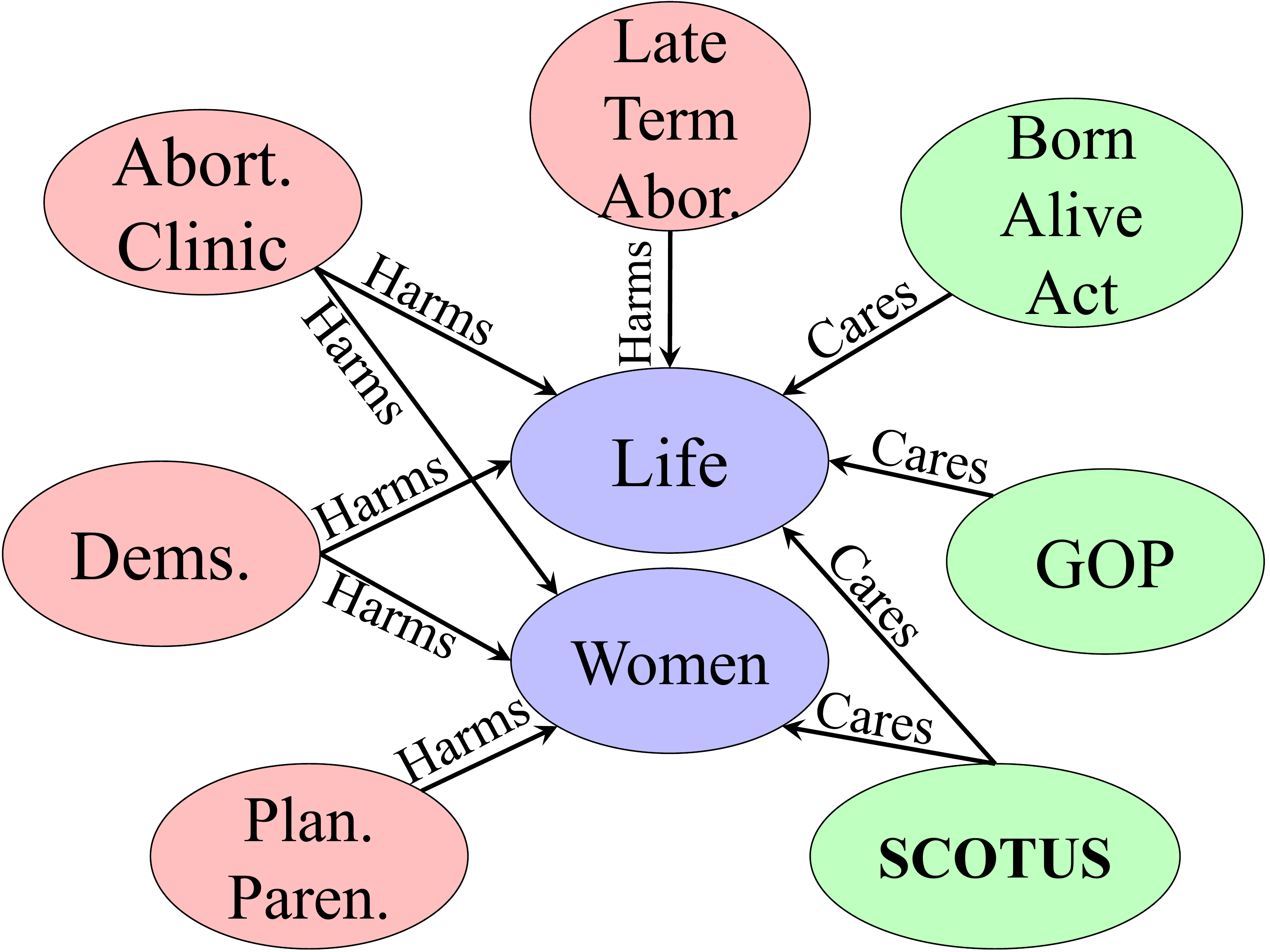}
        \caption{Entity relationship in right}
        \label{fig:entity-relation-right-abortion}
    \end{subfigure}
    \caption{Entity-relation graphs for moral foundation \textbf{Care/Harm}. Here, blue, red and green spheres represent target, harming and caring entities, respectively. A directed edge represents relationship between two entities.
    }
    \label{fig:entity-relation-abortion}
    \end{center}
\end{figure*}

We performed human evaluation on the predictions for this new data by randomly sampling $50$ tweets from each issue. This resulted in $91$ and $76$ (tweet, entity) pairs for Abortion and US Capitol, respectively. 
This procedure resulted in an accuracy of MF prediction of $88\%$ for each issue, and a role prediction accuracy of $75\%$ for Abortion, and $60.44\%$ for the US Capitol incident. 
We found that entities that appear less in the training data have low precision for the role prediction (See Table \ref{tab:train_freq}). 
Note that the US Capitol event was not observed during training, which makes it more challenging.   
For Abortion, we observed that Democrats mention the entity \textit{Women} most, and $84\%$ of the time the predicted MF role is target of care/harm or fairness/cheating, and it is never assigned a negative role (possibly because of constraint $c_3$). For Republicans, we observed the same pattern for the entity \textit{Life} (Stats in Appendix \ref{appendix:role-dist}). However, in a few cases (2.4\%) \textit{Life} is predicted as the entity ensuring fairness/purity/care, justified authority or being loyal. While these roles carry a positive sentiment, they are intuitively wrong predictions for \textit{Life}. We found out that for $34.21\%$ of such cases, there are multiple mentions of \textit{Life} in the same tweet. Given that constraint $c_2$ encourages different roles for different entities in a tweet, this can be the source of this error. Examples for these cases can be found in Appendix \ref{appendix:life-qualitative-eval}.

%% file: 6qualitative.tex
\begin{figure}[h]
    \centering
    \begin{subfigure}[t]{0.50\textwidth}
        \centering
        \includegraphics[width=0.85\textwidth]{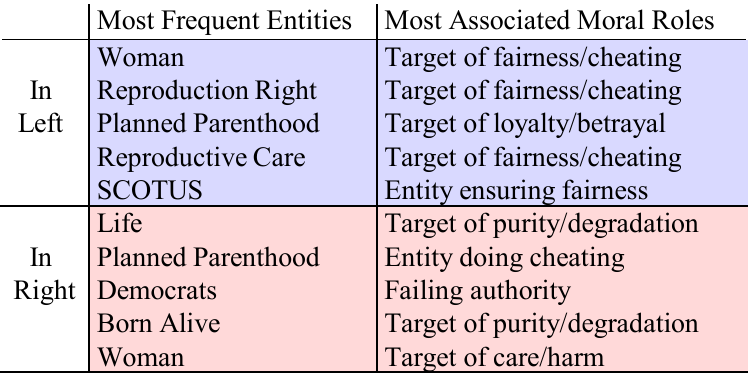}
        \caption{Most frequent entities \& most associated moral roles.}
        \label{fig:most_used_entities_abortion}
    \end{subfigure}%
    \\
    \centering
    \begin{subfigure}[t]{0.45\textwidth}
        \centering
        \includegraphics[width=0.7\textwidth]{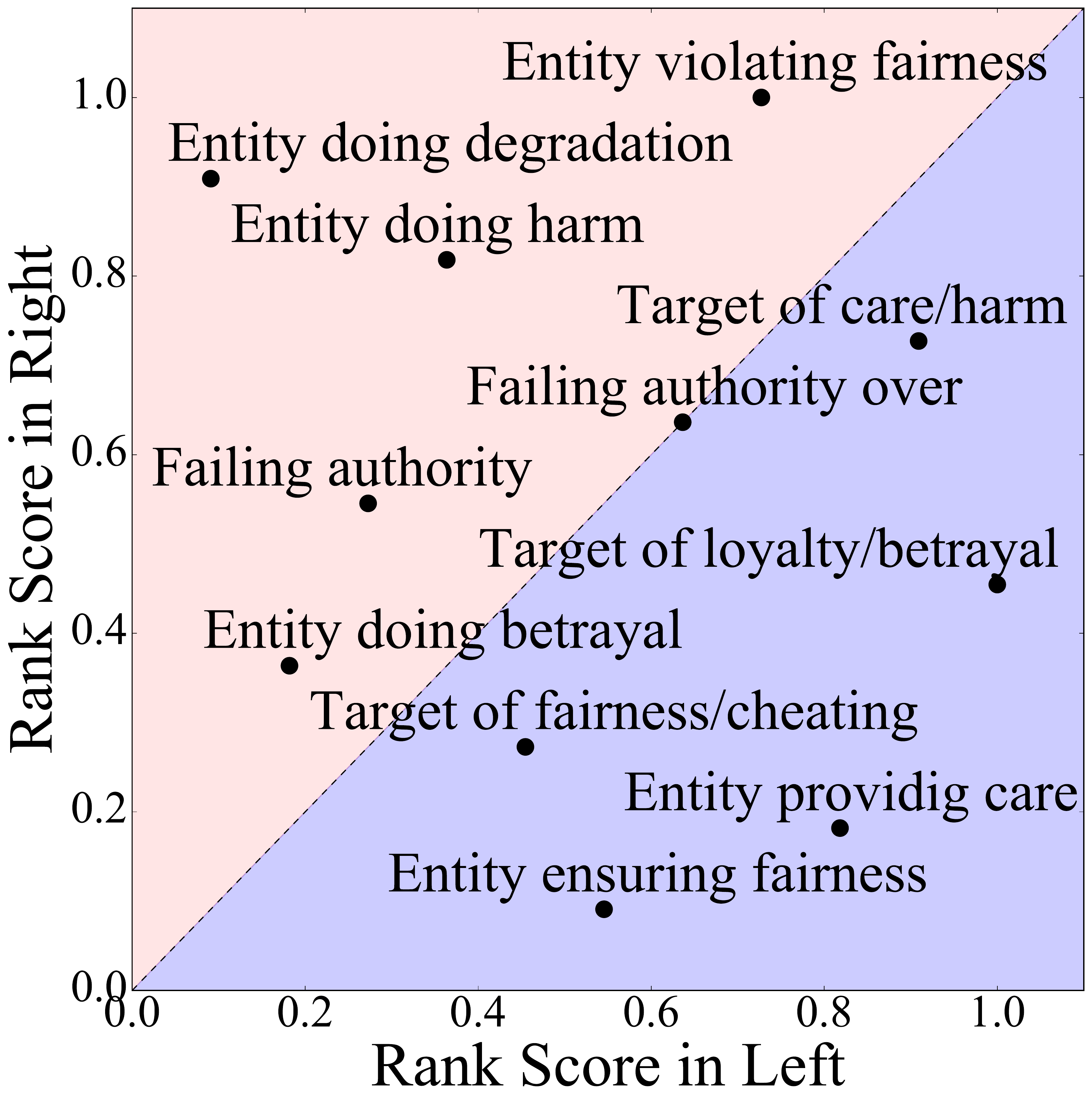}
        \caption{Sentiment towards \textbf{Planned Parenthood}. Normalized rank scores of MF roles based on usage, are plotted in (x, y)-axes. We discarded roles used <10 times.
        }
        \label{fig:polarization}
    \end{subfigure}%
    \caption{Polarization in sentiment towards entities.}
\end{figure}

\begin{table*}[h]
\begin{center}
\scalebox{0.66}{\begin{tabular}{>{\arraybackslash}m{4cm}||>{\arraybackslash}m{4.7cm}|>{\arraybackslash}m{5cm}||>{\arraybackslash}m{3.9cm}|>{\arraybackslash}m{4.4cm}} 
 \hline 
\textbf{\textsc{Entity Types}} & \multicolumn{2}{c}{\textsc{\textbf{Top Entities in Left}}} & \multicolumn{2}{c}{\textsc{\textbf{Top Entities in Right}}}\\ [0.5ex]
\cline{2-5}
&   \textsc{\textbf{Pre-Event}}   &   \textsc{\textbf{Post-Event}}   &   \textsc{\textbf{Pre-Event}}   &   \textsc{\textbf{Post-Event}}\\
 \hline
 \textsc{\textbf{Target of care}} & Citizens, Democracy, America & Capitol, Democracy, Police & America, Citizens & Capitol, America, Sicknick\\
 \hline
 \textsc{\textbf{Causing harm}} & Trump, Violence & Trump, Violence, Domest. terror. & - & Violence, Trump\\
 \hline
 \textsc{\textbf{Provide care}} & Congress, Biden, Democrats & Congress, Biden, Amendment & Congress, Trump & Police; Congress\\
 
 \hline
 \hline
 \textsc{\textbf{Justified authority}} & Congress, Pelosi, Democrats & Congress, Amendment, Pence & Congress & Congress, Pence\\
 \hline
 \textsc{\textbf{Justified auth. over}} & - & Biden, Harris & - & - \\
 \hline
 \textsc{\textbf{Failing authority}} & Trump, GOP & Trump, Impeachment, GOP & Democrats, Trump, GOP & Trump, Dems., Impeachment\\
 \hline
 \textsc{\textbf{Failing auth. over}} & Democracy, Biden, McConnell & Democracy, Capitol, Nation & Pelosi, Citizens, America & Nation, Pelosi, Biden\\
 \hline
 \end{tabular}}
\caption{Top-3 target entities per role pre and post the US Capitol event on January 6, 2021. Entity roles shown in this table are related to the moral foundations Care/Harm and Authority/Subversion.}
\label{tab:top3entities}
\end{center}
\end{table*}

\section{Analyzing Political Discussions}\label{sec:qualitative_eval}
In this section, we first characterize the political discussion on \textit{Abortion} using the predicted morality frames. Then, we analyze how an event impacts the moral sentiment towards entities by looking at the usage of MF roles before and after the \textit{2021 US Capitol Storming} for the different parties. 

\subsection{Characterizing Discussion on Abortion}

\textbf{Morality Frame Usage:} We found out that the left uses Fairness/Cheating the most, while the right uses Purity/Degradation. Care/Harm is the second most frequent for both parties (Appendix \ref{appendix:moral-foundation-usage}). To analyze MF role usage, we list the most frequent entities and their most frequent moral roles in Figure \ref{fig:most_used_entities_abortion}. The left portrays entities related to \textit{Reproduction Freedom} as the target of Fairness/Cheating. While on the right, the top target of Purity/Degradation is \textit{Life}. Both of them use \textit{Planned Parenthood} frequently, but their sentiment towards it differs. To further examine this, we plot \textit{Planned Parenthood}'s polarity graph in Figure \ref{fig:polarization}. It shows that parties express opposite sentiments towards \textit{Planned Parenthood}. These findings are consistent with known stances of democrats and republicans on this topic.


\textbf{Entity-Relation Graph:} We examine how the political discussion is framed by each party by looking at the sentiments expressed towards different entities, regardless of whether they use the same high level MF. We look at Care/Harm, which is frequently used by both parties, and take the two most used targets by each party. We then take the top three care providing and harming entities used in the same tweet as the target. We assign the most common role for each entity, and represent it in an entity-relation graph in Figure \ref{fig:entity-relation-left-abortion}-\ref{fig:entity-relation-right-abortion}.  We can see that both democrats and republicans express care for \textit{Women}, but the caring and harming entities vary highly across parties. For example, the left portrays \textit{Planned Parenthood} as the caring entity, while the right portrays it as the harming entity. This analysis shows that, while there is overlap in the MFs used, the moral roles of entities can highlight the differences between parties in politically polarized discussions at an \textbf{aggregate} level.

\subsection{Moral Response to US Capitol Storming}
To analyze how the moral sentiment towards entities changed after the storming of the US Capitol on January 6, 2021, we look at the sentiment towards entities before and after the event. We found that Authority/Subversion and Care/Harm were the two most used moral foundations after the incident for both parties (Appendix \ref{appendix:moral-foundation-usage}). In Table \ref{tab:top3entities}, we present the top three most frequent entities for role types under Care/Harm and Authority/Subversion, before and after the event. Entities appearing less than 15 times are omitted from this analysis. Our model predicted that, after the event, the left justified the authority of \textit{Mike Pence}, and \textit{violence} appeared as a harming entity even before the event occurred. On the right, \textit{Trump} shifted from an entity providing care prior to the event, to a harming entity after the event. We show some relevant tweets and their corresponding predictions in Table \ref{tab:qualitative-capitol-hill}. The entity-relation graph for each party after the event can can be found in Appendix \ref{appendix:entity-relation-ch}.

\begin{table}
\begin{center}
 \scalebox{0.75}{\begin{tabular}{>{\arraybackslash}m{10cm}}
 \toprule
\textbf{[Ideology-Period] (Predicted MF) Tweet Text}\\ 
 \toprule
 \makecell[l]{\textbf{[\textcolor{blue}{Left}-Pre-Event] (Care/Harm)} Zealotry. \textcolor{purple}{$[$}Trump\textcolor{purple}{$]_{DO-HARM}$}\\has seeded an anti-government fanaticism among his most fervent\\followers, threatening systematic \textcolor{purple}{$[$}violence\textcolor{purple}{$]_{DO-HARM}$} and the\\future of \textcolor{purple}{$[$}American democracy\textcolor{purple}{$]_{TARGET-CARE}$}.}\\
 \midrule
 \makecell[l]{\textbf{[\textcolor{blue}{Left}-Post-Event] (Authority/Subversion)} I'm calling on @VP\\\textcolor{purple}{$[$}Pence\textcolor{purple}{$]_{JUST-AUTH}$} to invoke the \textcolor{purple}{$[$}25th amendment\textcolor{purple}{$]_{JUST-AUTH}$}\\in order to immediately remove President \textcolor{purple}{$[$}Trump\textcolor{purple}{$]_{FAIL-AUTH}$}\\from office.}\\
 \midrule
 \makecell[l]{\textbf{[\textcolor{red}{Right}-Pre-Event] (Care/Harm)} Great news:\\\textcolor{purple}{$[$}@realDonaldTrump\textcolor{purple}{$]_{PROVIDE-CARE}$} just signed the \#GLRI Act\\into law - bipartisan legislation that will help protect \& preserve\\the Great Lakes.}\\
 \midrule
 \makecell[l]{\textbf{[\textcolor{red}{Right}-Post-Event] (Care/Harm)} President \textcolor{purple}{$[$}Trump's\textcolor{purple}{$]_{DO-HARM}$}\\incendiary rhetoric and false election fraud claims incited his\\supporters to \textcolor{purple}{$[$}violence\textcolor{purple}{$]_{DO-HARM}$}.}\\ 
 \bottomrule
 \end{tabular}}

\caption{Examples of moral roles prediction for entities related to the US Capitol event on January 6, 2021.}
\label{tab:qualitative-capitol-hill}
\end{center}

\end{table}

%% file: 7summary.tex
\section{Summary}
In this paper, we present the first study on Moral Foundations Theory at the entity level, by assigning moral roles to entities, and present a novel dataset of political tweets that is annotated for this purpose. We propose a relational model that predicts moral foundations and the moral roles of entities jointly, and show the effectiveness of modeling dependencies and contextualizing information for this task. Finally, we analyze political discussions in the US using these predictions, and show the usefulness of our proposed schema. In the future, we intend to study how morality frames and our relational framework can be applied in other settings, where contextualizing information is not observed.

\section*{Acknowledgements}\label{sec:acknowledgements} We thank Nikhil Mehta, Rajkumar Pujari, and the anonymous reviewers for their insightful comments. This work was partially supported by an NSF
CAREER award IIS-2048001.

%% file: 8ethical.tex
\clearpage
\section{Ethics Statement}
To the best of our knowledge no code of ethics was violated throughout the annotations and experiments done in this paper. We used human annotation for annotating an existing dataset with new labels. We adequately acknowledged the dataset and its various properties are explained thoroughly in the paper. While annotating, we respected the privacy rights of the crowd annotators and we didn't ask any personal details of the anonymous human annotators. They were informed that the task contains potentially sensitive political content. The crowd annotators were fairly compensated by rewards per annotation. We determined what is a fair amount of compensation by taking into consideration the feedback from the annotators and comparing our reward with other annotation tasks on the crowd-sourcing platform. 

The dataset presented is comprised of tweets, and for the reviewers, we only submitted a subset of the tweets with text. We will replace the tweet text with only tweet ids when publishing it publicly to respect the privacy policy of Twitter. We did a thorough qualitative and quantitative evaluation of our annotated dataset, presented in the paper. We reported all pre-processing steps, hyper-parameters, other technical details and will release our code and data for reproducibility. Due to space constraints, we moved some of the pre-processing steps, detailed hyper-parameter information, and additional results to the Appendix section. The results reported in this paper support our claims and we believe that they are reproducible. Any qualitative result we report is an outcome from a machine learning model and does not represent the authors' personal views, nor the official stances of the political parties analyzed. As we study text from humans to identify the moral sentiment, to draw conclusions, we rely on a machine learning model which is more interpretable than an end to end deep learning model.




%% file: appendix.tex
\appendix

\section{Data Collection}\label{appendix:dataset}

\paragraph{Identification and Annotation of Tweets with `Purity/Degradation':}\label{appendix:sanctity-annotation}
To collect more tweets on Purity/Degradation, we took more examples from the unlabeled segment of the dataset (93K tweets). Then we filtered out $619$ tweets from it based on lexicon matching with Moral Foundation Dictionary for Purity/Degradation. Then two of the authors of this paper individually went over the $619$ tweets and selected tweets having purity/degradation as the primary moral foundation in them. The two authors had agreement on $95\%$ of the cases. Then we combined the two lists from two authors and in case of a disagreement we resolved it by discussion. In this manner we found $44$ tweets on Purity/Degradation. Then we annotate these $44$ tweets with Purity/Degradation with $17$ Policy Frames present in them in the same manner. Two authors of this paper annotated the $44$ tweets for Policy Frames individually. They had an agreement on $47\%$ of the cases about the primary policy frame in a tweet. Most of the time they had a disagreement in the cases where there are more than $1$ policy frame present in them. The authors resolved any disagreement by discussion.

\vspace{5pt}
\paragraph{Full Dataset Statistics:}\label{appendix:full-data-stat}
The statistics of the full dataset can be found in Table \ref{tab:dataset_summary-full}.

\begin{table}[ht]
\begin{center}
 \scalebox{0.45}{\begin{tabular}{>{\arraybackslash}m{3.5cm}|>{\centering\arraybackslash}m{1.3cm}||>{\centering\arraybackslash}m{1cm}|>{\centering\arraybackslash}m{1cm}||>{\centering\arraybackslash}m{1cm}|>{\centering\arraybackslash}m{1cm}|>{\centering\arraybackslash}m{1cm}|>{\centering\arraybackslash}m{1cm}|>{\centering\arraybackslash}m{1cm}|>{\centering\arraybackslash}m{1cm}} 
 \hline 
\textsc{\textbf{Morals}} & \textsc{\textbf{\# of Tweets}} & \multicolumn{2}{c}{\textsc{\textbf{Ideology}}} & \multicolumn{6}{c}{\textsc{\textbf{Topic}}}\\ [0.5ex]
\cline{3-10}
&   & \textsc{Left}   &   \textsc{Right}   &   \textsc{ABO}   &   \textsc{ACA} &   \textsc{GUN} &   \textsc{IMM} &   \textsc{LGBT} &   \textsc{TER}\\
 \hline
 Care/Harm              & 589 & 378 & 211 & 30 & 142 & 221 & 31 & 11 & 154\\
 Fairness/Cheating      & 264 & 201 & 63  & 42 & 81  & 33  & 22 & 73 & 13\\
 Loyalty/Betrayal       & 231 & 167 & 64  & 15 & 20  & 92  & 28 & 24 & 52\\
 Authority/Subversion   & 471 & 200 & 271 & 33 & 177 & 76  & 99 & 19 & 67\\
 Purity/Degradation     & 44  & 13  & 31  & 21 & 3   & 8   & 6  & 2  & 4\\
 \hline 
 \textsc{\textbf{Total}}& 1599& 959 & 640 & 141& 423 &430  &186 &129 &290\\
 \hline
 \end{tabular}}
\caption{Dataset summary.}
\label{tab:dataset_summary-full}
\end{center}
\end{table}

\section{Data Annotation for Moral Roles}
\subsection{Questionnaire asked to the annotators for annotation of entity roles}\label{appendix:questionnaire-mturk-full}

The questionnaire asked to the annotators for all moral foundations can be found in Table \ref{tab:questionnaire-mturk}.

\begin{table*}[ht]
\begin{center}
\scalebox{0.65}{
\begin{tabular}{>{\arraybackslash}m{3.3cm}|>{\arraybackslash}m{5cm}|>{\arraybackslash}m{10cm}}
    \hline
    \textbf{\textsc{Moral}} & \textbf{\textsc{Entity Type}} & \textbf{\textsc{Question Asked to The Annotators}}\\ [0.5ex]
    \hline

    \multirow{3}{*}{Care/Harm}
    &Target of care/harm 
    & Which entity needs CARE, or is being HARMED?\\
    \cline{2-3}
    &Entity causing harm 
    & Which entity is causing the HARM?\\
    \cline{2-3}
    &Entity providing care 
    & Which entity is offering/providing the CARE?\\
    \hline
    
    \multirow{4}{*}{Fairness/Cheating}
    & N/A (additional question) 
    & Fairness or cheating on what?\\
    \cline{2-3}
    &Target of fairness/cheating 
    & Fairness for whom or who is being cheated?\\
    \cline{2-3}
    &Entity ensuring fairness 
    & Who or What is ensuring fairness or in charge of ensuring fairness?\\
    \cline{2-3}
    &Entity doing cheating 
    & Who or What is cheating or violating the fairness?\\
    \hline

    \multirow{5}{*}{Loyalty/Betrayal}
    & N/A (additional question) 
    & What are the phrases invoking LOYALTY?\\
    \cline{2-3}
    & N/A (additional question) 
    & What are the phrases invoking BETRAYAL?\\
    \cline{2-3}
    &Target of loyalty/betrayal 
    & LOYALTY or BETRAYAL to whom or what?\\
    \cline{2-3}
    &Entity being loyal 
    & Who or what is expressing LOYALTY?\\
    \cline{2-3}
    &Entity doing betrayal 
    & Who or what is doing BETRAYAL?\\
    \hline
    
    \multirow{5}{*}{Authority/Subversion}
    & N/A (additional question) 
    & LEADERSHIP or AUTHORITY on what issue or activity?\\
    \cline{2-3}
    &Justified authority 
    & Which LEADERSHIP or AUTHORITY is obeyed/praised/justified?\\
    \cline{2-3}
    &Justified authority over
    & If the LEADERSHIP or AUTHORITY is obeyed/praised/justified, then praised/obeyed by whom or justified over whom?\\
    \cline{2-3}
    &Failing authority
    & Which LEADERSHIP or AUTHORITY is disobeyed or failing or criticized?\\
    \cline{2-3}
    &Failing authority over
    & If the LEADERSHIP or AUTHORITY is disobeyed or failing or criticized, then failing to lead whom or disobeyed/criticized by whom?\\
    \hline
    
    \multirow{3}{*}{Purity/Degradation}
    &Target of purity/degradation 
    & What or who is SACRED, or subject to degradation?\\
    \cline{2-3}
    &Entity preserving purity 
    & Who is ensuring or preserving the sanctity?\\
    \cline{2-3}
    &Entity causing degradation 
    & Who is violating the sanctity or who is doing degradation or who is the target of disgust?\\
    \hline

\end{tabular}
}
\caption{Questionnaire asked to Amazon Mechanical Turk annotators for annotation of entities.}
\label{tab:questionnaire-mturk}
\end{center}
\end{table*}

\subsection{Calculation of Partisanship and Most Frequent Entities by Entity Role}\label{appendix:most-frequent-entities-by-entity-role}
To determine the partisanship of the elements - (1) moral foundations, (2) (moral foundation role: entity), we use z-score measure of these elements in the two political ideologies (left, right). We calculate the z-score to evaluate - whether two groups (e.g., left and right) differ significantly on some single characteristic. In our case the characteristics are any element of type (1) or type (2) as described above. A positive z-score means it's left-partisan and negative score means right-partisan. 

Most frequent entities per moral role can be found in Table \ref{tab:entity_freq}.

\begin{table*}[ht]
\begin{center}
\scalebox{0.6}{
\begin{tabular}{>{\arraybackslash}m{3.3cm}|>{\centering\arraybackslash}m{3.5cm}|>{\arraybackslash}m{9cm}|>{\arraybackslash}m{9cm}}
    \hline
    \textbf{\textsc{Moral}} & \textbf{\textsc{Entity Type}} & \textbf{\textsc{Most Frequent Entity in Left}} & \textbf{\textsc{Most Frequent Entity in Right}}\\ [0.5ex]
    \hline
    \multirow{3}{*}{Care/Harm}
    &Target of care/harm 
    & 20 million Americans; our families; woman; innocent people; \#domesticviolence victims 
    & law-abiding Americans; victims and their families; small businesses; patients; Paris\\
    \cline{2-4}
    &Entity causing harm 
    & gun show loopholes; gun violence; terrorist attack; mass-shootings; suspected terrorists 
    & Radical Islamic terrorists; \#Obamacare mandates; Brussels attacks; \#ISIS; ISIL-Inspired Attacks\\
    \cline{2-4}
    &Entity providing care 
    & gun show loophole bills; Affordable Care Act; \#ImmigrationReform; Democrats; commonsense gun legislation 
    & @RepHalRogers: Bill; @HouseGOP; Senate; @WaysandMeansGOP; HR 240\\
    \hline
    
    \multirow{3}{*}{Fairness/Cheating}
    &Target of fairness/cheating 
    & woman, \#LGBT community; all Americans; \#FightForFamilies; other vulnerable people
    & the American people; small businesses; people; religious minorities in Syria and Iraq\\
    \cline{2-4}
    &Entity ensuring fairness 
    & \#SCOTUS decision; congress; bill to expand access; the DREAM Act; Equality Act 
    & Senate; House; @RepHalRogers: Bill; House GOP; Supreme Court\'s \#HobbyLobby ruling\\
    \cline{2-4}
    &Entity doing cheating 
    & anti-\#LGBT laws; employer; HB 2; \#HobbyLobby decision; Political attacks 
    & \#Obamacare legislation; fake ISIS passports; Planned Parenthood; the Pakistani Gov; enforcement loopholes\\
    \hline
    
    \multirow{3}{*}{Loyalty/Betrayal}
    &Target of loyalty/betrayal 
    & \#LGBT communities; gun safety measures; victims of \#Orlando; women men and families; \#StandwithPP 
    & Paris terror attacks; senators; Israel; The American people; Syrian and Iraqi refugees\\
    \cline{2-4}
    &Entity being loyal 
    & @SenWarren; @RepAdams; My colleagues; @SenateDems; House Democrats 
    & @SenatorIsakson; @RepHalRogers\\
    \cline{2-4}
    &Entity doing betrayal 
    & @HouseGOP extremist Members!; terrorists; The community of nations; @NRA 
    & House\\
    \hline
    
    \multirow{4}{*}{Authority/Subversion}
    &Justified authority 
    & POTUS; SCOTUS; President Obama\'s; Senate; @HouseDems 
    & @HouseGOP; \#Senate; \#SCOTUS; Congress; Republicans\\
    \cline{2-4}
    &Justified authority over
    & Americans; 180 House Dems; nation; people 
    & @SenateMajLdr; @RepHalRogers; \#American; @SenateMajLdr; Inhofe\\
    \cline{2-4}
    &Failing authority
    & \#HouseGOP; Congress; Republicans; SCOTUS; @SpeakerRyan 
    & President Obama; POTUS; \#Obamacare; @SCOTUS; @SecBurwell\'s\\
    \cline{2-4}
    &Failing authority over
    & Americans; @repjohnlewis; family; @SenFeinstein\'s; women; Sen Dems 
    & Americans; @SenateMajLdr; @HouseAppropsGOP @RepHalRogers; @SenateMajLdr McConnell; @SpeakerRyan\\
    \hline
    
    \multirow{3}{*}{Purity/Degradation}
    &Target of purity/degradation 
    & immigration; women 
    & fetal body parts; lives of the unborn; baby girls\\
    \cline{2-4}
    &Entity preserving purity 
    & N/A (no ngrams found that occurs more than 2 times.) 
    & @SenDanCoats; \#MarchforLife\\
    \cline{2-4}
    &Entity causing degradation 
    & Donald Trump; Charleston church killings 
    & Planned Parenthood; abortion providers; Radical Islamic terrorists\\
    \hline

\end{tabular}
}
\caption{Most frequent entities by ideology and by moral role type. For each ideology the most frequent list was generated by taking the most common stemmed ngrams (n = 1 to 5) in the identified entities by the annotators. One representative entity from each ngram group is presented in this table. Inclusive ngrams were merged together. For example: `law abiding citizen' is merged with `law abiding'. }
\label{tab:entity_freq}
\end{center}
\end{table*}

\subsection{Expressivity of bias of Moral Roles}
To examine how well moral roles account for political standpoints when compared to moral foundations, we use the moral foundations (MF) and (moral foundation role, entity) (MFR) as one hot encoded features to classify the ideology of the tweet (left/right). The results are shown in Tab. \ref{tab:classification-one-hot}. Moral roles classify the ideology reasonably well compared to MF and BoW features, which proves the usefulness of the moral roles for capturing political perspectives. 

 \begin{table}[h!]
 \begin{center}
 \scalebox{0.65}{
 \begin{tabular}{>{\raggedright\arraybackslash}m{4.5cm}|>{\centering\arraybackslash}m{2cm}|>{\centering\arraybackslash}m{2cm}}
     \hline
     \textbf{One-hot Encoded Features} & \# of Features & \textbf{Macro F1} \\[0.5ex]
     \hline
     Moral Foundation (MF)   &  5       &   0.62\\
     Moral Roles (MFR)       &  2021    &   0.77\\
     MF+MFR                  &  2026    &   0.79\\
     Bag of Words (BoW)      &  2478    &   0.85\\
     BoW+MF                  &  2483    &   0.85\\
     BoW+MFR                 &  4499    &   0.87\\
     BoW+MF+MFR              &  4504    &   0.86\\
     \hline
 \end{tabular}
 }
 \caption{Predicting ideology of tweet using Logistic Regression (3-fold CV).}
 \label{tab:classification-one-hot}
 \end{center}
 \end{table}

\section{Modeling}

\subsection{Polarity of Moral Roles}\label{appendix:moral-role-polarity}

\textbf{Moral Roles with positive polarity:} Target of care/harm, Entity providing care, Target of fairness/cheating, Entity ensuring fairness, Target of loyalty/betrayal, Entity being loyal, Justified authority, Justified authority over, Failing authority over, Target of purity/degradation, Entity preserving purity.

\noindent\textbf{Moral Roles with negative polarity:} Entity causing harm, Entity doing cheating, Entity doing betrayal, Failing authority, Entity causing degradation.

\subsection{Relational Learning Frameworks}\label{app:reln_models}
\subsubsection{Probabilistic Soft Logic}

PSL models are specified using weighted horn clauses, which are compiled into a Hinge-Loss Markov Random Field, a class of undirected probabilistic graphical model. In HL-MRFs, a probability distribution is defined over continuous values in the range of [0, 1], and dependencies among them are modeled using linear and
quadratic hinge functions. This way, they define a probability density function:

\begin{equation}\small
\begin{split}
P(\bm{Y}|\bm{X}) &= \frac{1}{Z} \exp{(-\sum^M_{r=1} w_r \psi_r(\bm{X},\bm{Y}))}
\end{split}\label{eq:psl_density}
\end{equation}

\noindent where $w_r$ is the rule weight, $Z$ is a normalization constant and $\psi_r(\bm{Y},\bm{X}) = \max{\{l_r(\bm{X},\bm{Y}), 0\}}^{\rho_r}$ is the hinge-loss potential corresponding to the instantiation of rule $r$, represented by a linear function $l_r$ of $\bm{X}$ and $\bm{Y}$, and an optional exponent $\rho_r \in \{1, 2\}$. Inference in PSL is performed by finding a MAP estimate of the random variables $\bm{Y}$ given evidence $\bm{X}$, this is done by maximizing the density function in Eq. \ref{eq:psl_density} as $\argmax_{\bm{Y}} P(\bm{Y}|\bm{X})$. To solve this, they use Alternating Direction Method of Multipliers (ADMM), an efficient convex optimization procedure. 

Weights can be learned through maximum likelihood estimation by using the structured perceptron algorithm. The partial derivative of the log of the likelihood function in Eq. \ref{eq:psl_density} above with respect to a parameter $\bm{W}_r$ is:

\begin{equation}\small
\begin{split}
\frac{\partial\log P(\bm{Y}|\bm{X})}{\partial \bm{W}_r} &= \E_{\bm{W}} [ \psi_r(\bm{X},\bm{Y})] - \psi_r(\bm{X},\bm{Y})
\end{split}
\end{equation}

\noindent where $\E_{\bm{W}}$ is the expectation under the distribution defined by $\bm{W}$. Given that computing this expectation is intractable, they approximate it by taking the values in the MAP state. This approximation makes this learning approach a structured variant of the voted perceptron. Note that alternative estimations are also supported. More details can be found in the original paper \cite{bach:jmlr17}.

\subsubsection{DRaiL}

Rules in DRaiL can be \textit{weighted} (i.e. classifiers, soft constraints) or \textit{unweighted} (i.e. hard constraints). The collection of all rules represents the global decision. Rules are transformed into linear inequalities, corresponding to their disjunctive form, and MAP inference is then defined as an integer linear program:
\begin{equation}\small
\begin{split}
\argmax_{\bm{y}\in \{0,1\}^n} P(\bm{y}|\bm{x}) \equiv &\argmax_{\bm{y}\in \{0,1\}^n} \sum_{\psi_{r,t} \in \Psi} w_{r}~ \psi_{r}(\bm{x_r,y_r})\\
&\textit{s.t.}~c(\bm{x_c,y_c})\leq0 ; \;\;\forall c \in C
\end{split}\label{eq:inference}
\end{equation}
\noindent Where each rule grounding $r$, generated from template $t$, with input features $\bm{x_r}$ and predicted variables $\bm{y_r}$ defines the potential $\psi_{r}(\bm{x_r,y_r})$, added to the linear program with a weight $w_{r}$. DRaiL implements both exact and approximate inference to solve the MAP problem, in the latter case, the AD$^3$ algorithm is used \cite{Martins_ad3}.

In DRaiL, weights $w_{r}$ are learned using neural networks defined over parameter set $\bm{\theta}$. Parameters can be learned \textit{locally}, by training each rule independently, or \textit{globally}, by using inference to ensure that the scoring functions for all rules result in a globally consistent decision. To train global models using large-margin estimation, DRaiL uses the structured hinge loss:
\begin{gather*}
 \max_{\hat{\textbf{y}} \in Y}(\Delta(\hat{\textbf{y}},\textbf{y}) + \sum_{\psi_r \in \Psi} \Phi_t(\bm{x_r,\hat{y}_r}; \theta^{t})) 
-\sum_{\psi_r \in \Psi} \Phi_t(\bm{x_r,y_r}; \theta^{t})
\label{eq:structured_hinge_loss}
\end{gather*}

\noindent Where $\Phi_t$ represents the neural network associated with rule template $t$, and parameter set $\theta^t$. Here, $\textbf{y}$ corresponds to the gold assignments, and $\hat{\textbf{y}}$ corresponds to the prediction resulting from the MAP inference defined in Eq. \ref{eq:inference}. Note that alternative estimations are also supported. More details can be found in the original paper~\cite{pacheco-goldwasser-2021-modeling}.

\section{Experimental Evaluation}\label{appendix:experimental-eval}

\subsection{Task-Adaptive Pretraining}\label{appendix:pretraining}

We do task-adaptive pretraining for BERT~\cite{gururangan-etal-2020-dont}, and fine-tune it on a large number of unlabeled tweets\footnote{Collected From:\\ https://github.com/alexlitel/congresstweets}. To select unlabeled tweets, we build a topic-specific lexicon of n-grams (n$\leq$ 5) from our training dataset based on Pointwise Mutual Information (PMI) scores \cite{church-hanks-1990-word}. 
%
Namely, for an ngram $w$ we calculate the point-wise mutual information (PMI) with label $l$ (e.g. topic), $I(w, l)$ using the following formula.
\begin{align*}
     I(w,l)=\operatorname{log}\frac{P(w|l)}{P(w)}
\end{align*}{}

Where $P(w|l)$ is computed by taking all tweets with label $l$ and computing $\frac{count(w)}{count(allwords)}$. Similarly, $P(w)$ is computed by counting ngram $w$ over the set of tweets with any label. To construct the lexicon, we rank ngrams for each label based on their PMI scores. 

We explore three pretraining objectives, described below. In all cases, models were initialized using BERT \cite{devlin-etal-2019-bert}. 

\textbf{Masked Language Modeling:} We randomly mask some of the tokens from the input, and predict the original vocabulary id of the masked word based on its context \cite{devlin-etal-2019-bert}. 

\textbf{Whole Word Masking}: Instead of masking randomly selected tokens, which may be sub-segments of words, we mask randomly selected words. 

\textbf{Moral Foundations Dictionary:} We create a lexicon for each Moral Foundation from the dataset by \citet{johnson-goldwasser-2018-classification} using the same PMI formula described above. We use the normalized PMI scores as a weight for each unigram, and assign a weight of 1 to unigrams in the Moral Foundation Dictionary (MFD)\cite{DVN/SJTRBI_2009}. We score a tweet by summing the scores of words matching the lexicon. We take the highest scoring moral foundation for each tweet, and fine-tune a moral foundation classifier using this weakly annotated data. 

We evaluate these objectives by performing the pre-training stage on the unlabeled data, and fine-tuning the encoder for our base task of leveraging only text to predict moral foundations and entity roles. 
Results can be seen in Tab. \ref{tab:tapt}. 

\begin{table}[h!]
    \centering
    \resizebox{\columnwidth}{!}{%
    \begin{tabular}{lcccc}
    \toprule
        \multirow{2}{*}{\textbf{\textsc{Objective}}} & \multicolumn{2}{c}{\textbf{\textsc{Macro}}} & \multicolumn{2}{c}{\textbf{\textsc{Weighted}}} \\
        ~ & \textsc{Role} & \textsc{MF} & \textsc{Role} & \textsc{MF} \\
        \midrule
        BERT-base-uncased & 49.32 & 59.99 & 57.37 & 62.17 \\
        Masked LM & 53.49 & 63.90 & 61.51 & 67.45 \\
        Whole Word Masking & \textbf{54.73} & \textbf{66.44} & \textbf{62.18} & \textbf{68.29} \\
        MF Dictionary &  47.92 & 63.70 & 54.93 & 65.81 \\
    \bottomrule
    \end{tabular}
    }
    \caption{Task-Adaptive Pretraining (F1 Scores)}
    \label{tab:tapt}
\end{table}

\subsection{Details About the Baselines}
\noindent\textbf{Lexicon Matching:} We label a tweet with the moral foundation with maximum score based on lexicon matching. We use the Moral Foundation lexicons created in Appendix \ref{appendix:pretraining}. If there is no lexicon matching for a tweet, we assign a moral foundation label to it randomly. We experiment with combining and not combining the Moral Foundations Dictionary (MFD) \cite{DVN/SJTRBI_2009}. 

\noindent\textbf{Sequence Tagging:}
We use a bidirectional LSTM with a CRF layer on top for tagging each entity a tweet with a moral role label. We run two LSTMs in forward and reverse direction of a tweet and concatenate the hidden states (50d) of two directions at each time step to get an embedding (100d) of the token. Given that entity spans are known, we use the last token in each entity as the entity embedding. This embedding is then used for the CRF layer. 

\noindent\textbf{End-to-end Classifiers:} For the classification of moral foundations using BiLSTM, we run two opposite directional LSTMs over the GloVe word embeddings~\cite{pennington2014glove} of all tokens of a tweet, concatenate the hidden states (150d) of both LSTMs to get the embedding of a token (300d), then average the embeddings of all tokens to get a final embedding of a tweet. Then we use this embedding to classify the tweet in the moral foundation classes using a fully connected layer that maps the embedding to a moral foundation class. For moral foundation role classification using BiLSTM, we repeat the same process for an entity text to get its representation using BiLSTM. Then we concatenate the tweet representation and the entity representation and pass it through a hidden layer to get a representation of size 300. Then we use this representation for classification of moral foundation roles using a fully connected layer that maps the representation to the moral foundation role classes. For BERT based models, we use a classifier on top of the \texttt{[CLS]} representation. For role classification, we pass an input of the form \texttt{[CLS] [tweet] [SEP] [entity]}. We use the default parameters of the BERT-base-uncased huggingface implementation. 

\noindent\textbf{Multitasking Based:}  We define a single BERT encoder, and a single ideology and topic embedding that is shared across the two tasks. The three representations are concatenated and task-specific classifiers are used on top of them. Then, the loss functions are added as $L = \lambda_1 L_{\mathtt{MF}} + \lambda_2 L_{\mathtt{Role}}$. We set $\lambda_1 = \lambda_2 = 1$. For topic and ideology embeddings, we use feed-forward computations with 100 hidden layers and ReLU activations. For BERT we use the same configuration as the end-to-end classifiers. 

\subsection{Hyper-parameter Tuning and Validation Set Performance}\label{app:hyper}

For the underlying BERT, we use the default parameters of the hugging face implementation\footnote{\url{https://github.com/huggingface/transformers}}. Other parameters can be observed in Table \ref{tab:hyperparams-validation-performance} (top). The bottom part of Table \ref{tab:hyperparams-validation-performance} shows the validation performance during the learning of the best performing model.

\begin{table}[t!]
    \centering
    \resizebox{\columnwidth}{!}{%
    \begin{tabular}{llcc}
    \toprule
       Task & Param & Search Space & Selected Value \\
       \midrule
       Local  & Learning Rate & 5e-5, 2e-5, 1e-5, 1e-6  & 2e-5 \\
        (Base) & Batch size & 64, 32 & 32 \\
       ~ & Patience & 3, 5, 10 & 10 \\
       ~ & Optimizer & SGD,Adam,AdamW &  AdamW \\
       ~ & Hidden Units & - & 100 \\
       ~ & Non-linearity & -  & ReLU \\
       \midrule
       Local  & Learning Rate & 1e-3, 5e-3, 5e-2, 1e-2 & 5e-3 \\
        (Soft  & Batch size &  64, 32 & 32 \\
       Constr.) & Patience & 5, 10, 20 & 20    \\
       ~ & Optimizer & SGD,Adam,AdamW &  AdamW \\
       ~ & Hidden Units & - & 100 \\
       ~ & Non-linearity & - & ReLU \\
      \midrule
       DRail & Learning Rate & 5e-5, 2e-5, 1e-5, 1e-6 & 1e-6 \\
       Global    & Batch size & - & Full instance \\
       ~ & Patience & 3, 5, 10 & 10 \\
       ~ & Optimizer & SGD,Adam,AdamW &  AdamW \\
       ~ & Hidden Units & - & 100 \\
        ~ & Non-linearity & -  & ReLU \\
       \bottomrule
    \end{tabular}}
    
    \resizebox{\columnwidth}{!}{%
    \begin{tabular}{llcccc}
    \toprule
    \multirow{2}{*}{\textbf{\textsc{Group}}} & \multirow{2}{*}{\textbf{\textsc{Model}}} & \multicolumn{2}{c}{\textbf{\textsc{Macro}}} & \multicolumn{2}{c}{\textbf{\textsc{Weighted}}} \\
        ~ & ~ & \textsc{Role} & \textsc{MF} & \textsc{Role} & \textsc{MF} \\
        \midrule
    \multirow{3}{*}{Simple} & BERT  & 52.37 & 60.38 & 63.26 & 67.57 \\
           & + Ideo + Issue & 52.52  & 60.31 & 64.02 & 66.58  \\
           & Combined & 53.34  & 59.84 & 64.65 &  67.29 \\
    \midrule
    \multirow{2}{*}{Struct.} 
    & DRaiL Local   & 51.71 & 64.02 & 63.99 & 71.37  \\
    & DRail Global     & \textbf{53.23}  & \textbf{65.46} & \textbf{65.50} & \textbf{72.39}  \\
    \midrule
    \multirow{2}{*}{Skyline} & DRaiL Global & \multirow{2}{*}{76.85} & \multirow{2}{*}{-} & \multirow{2}{*}{86.27} & \multirow{2}{*}{-} \\
    
    & (Fixed MF) \\
    \bottomrule
    \end{tabular}
    }

    \caption{{\small Hyper-parameter tuning (top) and validation set performance on the best model that combines all rules and constraints in DRaiL (bottom).}}
\label{tab:hyperparams-validation-performance}
\end{table}

\subsection{Results per Class}\label{app:per_class}

The per class classification results can be found in the Table \ref{tab:role_mf_per_class}.

\begin{table}[h]
    \centering
    \scalebox{0.6}{
    \begin{tabular}{lcccc}
    \toprule 
    \textbf{\textsc{MF}} & \textbf{\textsc{Pre.}} & \textbf{\textsc{Rec.}} & \textbf{\textsc{F1}} & \textbf{\textsc{Sup.}}  \\
    \midrule
     \textsc{Auth/Subv}    &  84.64 & 78.31  & 81.35 & 415 \\
     \textsc{Care/Harm}   & 72.10 & 84.30 & 77.72 & 567 \\
     \textsc{Fair/Cheat} & 70.71  & 56.91 & 63.06  & 246\\
     \textsc{Loyal/Betray} & 66.18  & 63.68  & 64.90 & 212\\
     \textsc{Purity/Degrad} & 84.85 & 66.67 & 74.67  & 42 \\
     \midrule 
     Macro Avg. & \textbf{75.69}  & \textbf{69.97} & \textbf{72.34} & 1482\\
     Weight Avg. & \textbf{74.89} & \textbf{74.63} & \textbf{74.39} & \\
     \bottomrule
    \end{tabular}
    }
    \scalebox{0.6}{
    \begin{tabular}{llccccc}
    \toprule
        \textbf{\textsc{MF}} & \textbf{\textsc{Role}} & \textbf{\textsc{Pre.}} & \textbf{\textsc{Rec.}} & \textbf{\textsc{F1}} & \textbf{\textsc{F1-Sky}} & \textbf{\textsc{Sup.}}  \\
    \midrule
         & Justified & 56.43 & 50.64 & 53.38 & 67.09 & 156 \\
       \textbf{\textsc{Auth/}} & Justified Over & 47.69  &  46.27  & 46.97 & 64.52 & 67 \\
       \textbf{\textsc{Subv}} & Fail & 71.96 & 71.43  & 71.69   & 80.90 &   273 \\
       ~ & Fail Over & 67.76   & 65.91 & 66.82 & 80.56 &       220\\
     \midrule
     \textbf{\textsc{Care/}} & Target & 67.42   & 78.01 & 72.33 & 92.72 & 382\\
     \textbf{\textsc{Harm}} & Cause Harm & 82.41  & 8159  & 82.00  & 92.70 &  402\\
     ~ & Provide Care & 57.11  & 74.23 & 64.56 & 91.06 &      357\\
     \midrule
    \textbf{\textsc{Fair/}} & Target & 68.18 & 59.66 & 63.64 & 92.01 & 176\\
    \textbf{\textsc{Cheat}} & Ensure Fair & 67.22  & 54.02 & 59.90 & 91.65 &       224\\
     ~ & Do Cheat & 64.62 & 43.75  & 52.17 & 83.87 & 96\\
     \midrule
    \textbf{\textsc{Loyal/}} & Target & 55.47 & 54.87 & 55.17 & 77.96   & 277\\
    \textbf{\textsc{Betray}} & Be Loyal & 59.28 & 56.25 & 57.73 & 74.64 & 176\\
     ~ & Do Betray & 17.65  & 16.22 & 16.90 & 32.50 & 37\\
     \midrule
     \textbf{\textsc{Purity/}}  & Target & 60.00 & 56.76  & 58.33  & 81.01 & 37\\
     \textbf{\textsc{Degrad}} & Preserve Purity & 86.67 & 46.43 & 60.47 & 83.02 &    28\\
     ~ & Cause Degrad & 77.78  & 56.76  & 65.62 & 83.33 & 37\\
     \midrule
    & Macro Avg. & \textbf{62.98} & \textbf{57.05} & \textbf{59.23} & 79.35 & 2945\\
    & Weight Avg. & \textbf{65.56}  & \textbf{65.23}  & \textbf{64.98}  & 84.52 & \\
     \bottomrule
     
    \end{tabular}
    }
    
    \caption{Moral foundation classification results per class (top) and role classification results per class (bottom).}
    \label{tab:role_mf_per_class}
\end{table}

\subsection{Run-time Analysis}

All experiments were run on a 4 core Intel(R) Core(TM) i5-7400 CPU @ 3.00GHz machine with 64GB RAM and an NVIDIA GeForce GTX 1080 Ti 11GB GDDR5X GPU. Runtimes for our models can be found in Table \ref{tab:runtime}

\begin{table}[h]
    \centering
    \scalebox{0.7}{
    \begin{tabular}{lrcr}
    \toprule
       Task & sec p/Epoch & epochs p/It  & sec  p/It  \\
       \midrule
       $r_1$ local & 9.588 & 18 & 177.037\\
       $r_2$ local & 4.002 & 15 & 64.106 \\
       $r_3$ local & 9.762 & 27 & 269.350 \\
       $r_4$ local & 4.286 & 17 & 74.541 \\
       $c_2$ local & 0.138 & 68 & 144.631 \\
       Global learn & 49.259 & 25 & 1268.615 \\
       Global predict & - & - & 7.725 \\
    \bottomrule
    \end{tabular}}
    \caption{Average runtimes for 3 fold cross-validation}
    \label{tab:runtime}
\end{table}

\subsection{Entity Groups}\label{appendix:entity-groups}
\subsubsection{Entity Groups For Abortion}
\textbf{Brett Kavanaugh:} brett kavanaugh, kavanaugh, stop kavanaugh\\
\textbf{Roe v Wade:} roe v wade, commit roe, protect roe\\
\textbf{Planned Parenthood:} plan parenthood, stand pp, pp, ppfa, ppact\\
\textbf{Affordable Care Act:} aca, affordable health care\\
\textbf{Title X:} title x, family planning, protect x\\
\textbf{Gag Rule:} gag rule, global gag rule, domestic gag rule\\
\textbf{Democrats:} democrat, dem, house democrat\\
\textbf{Republicans:} republican, house gop, senate gop, gop, gop leader\\
\textbf{Trump care:} trump care\\
\textbf{Woman:} woman\\
\textbf{Reproductive Right:} reproductive right, woman reproductive right, reproductive freedom, reproductive justice, woman reproductive freedom\\
\textbf{Reproductive Health:} reproductive health, woman reproductive health, reproductive health care, reproductive care, reproductive healthcare, reproductive health service, comprehensive reproductive health care, abortion care\\
\textbf{SCOTUS:} scotus, save scotus, supreme court, supreme court justice, supreme court decision\\
\textbf{Life:} human life, innocent life, stand life, unborn child, unborn child protection, unborn baby, unborn, baby\\
\textbf{NRLC:} nrlc\\
\textbf{NARAL:} naral\\
\textbf{Born Alive:} bear alive abortion, bear alive\\
\textbf{Late Term Abortion:} late term abortion\\
\textbf{Late Term Abortion Ban:} week abortion ban\\
\textbf{Born Alive Act:} bear alive act\\
\textbf{Abortion Provider:} abortion provider, abortion clinic, abortion industry\\
\textbf{Hyde Amendment:} hyde, hyde amendment, bold end hyde\\
\textbf{Healthcare Decision:} health care decision, healthcare decision\\
\textbf{Medicaid:} medicaid\\
\textbf{Medicare:} medicare

\subsubsection{Entity Groups for the 2021 US Capitol Hill Storming Event}
\textbf{Congress:} congress, th congress\\
\textbf{POTUS:} potus, president\\
\textbf{Donald Trump:} trump, donald trump, real donald trump, president real donald trump\\
\textbf{America:} america\\
\textbf{American People:} american people\\
\textbf{Democracy:} american democracy, democracy\\
\textbf{Joe Biden:} joe biden, biden, president elect\\
\textbf{Amendment:} amendment, th amendment\\
\textbf{Brian Sicknick:} brian sicknick, sicknick\\
\textbf{Nancy Pelosi:} pelosi, speaker pelosi, nancy pelosi\\
\textbf{Jamie Raskin:} raskin\\
\textbf{Capitol:} capitol, capitol building, capitol hill, nation capitol\\
\textbf{Impeachment:} impeachment, impeach president\\
\textbf{Kamala Harris:} kamala harris, vice president elect\\
\textbf{Capitol Police:} capitol police, police officer, law enforcement, law enforcement officer\\
\textbf{Mike Pence:} pence, vp pence, mike pence\\
\textbf{Mitch McConnell:} mitch mcconnell, mcconnell\\
\textbf{GOP:} house gop, gop leader, gop, republican\\
\textbf{Domestic Terrorism:} domestic terrorist, domestic terrorism\\
\textbf{Nation:} nation\\
\textbf{National Security:} national security, national guard\\
\textbf{Democrats:} dem, democrat, house democrat\\
\textbf{Violence:} violence, violent insurrection, violent attack, violent mob\\
\textbf{White Supremacist:} white supremacist\\
\textbf{Fair Election:} fair election

\subsection{Human Evaluation on Test Data}\label{appendix:human-evaluation-test-data}
\paragraph{Model Prediction Validation} We trained our model with all of our labeled data and used it to predict the moral foundations and entity roles of (tweet, entity) pairs in the new set. The validation set (randomly selected from train set) weighted F1 scores were $72.20\%$ and $64.59\%$ for moral foundations and roles, respectively. We validate our model's prediction on the unseen dataset using human evaluation. We randomly sampled 50 tweets from each of the two test sets. This resulted in $91$ and $76$ (tweet, entity) pairs for Abortion and US Capitol, respectively. Note that one tweet may have $>1$ entities. Then, we presented the predictions of moral foundations and entity roles to two graduate students and asked them if the prediction is correct or not. We found the Cohen's Kappa \cite{cohen1960coefficient} score between the annotators to be $0.50$ (moderate agreement) and $0.64$ (substantial agreement) in case of the moral foundations and entity roles, respectively. In case of a disagreement, we asked a third grad student to break the tie. The accuracy of the model for moral foundations was $88\%$ for each topic, while for roles it was  $75\%$ and $60.44\%$, for Abortion and US Capitol, respectively.

\subsection{Distribution of MF Roles Assigned by the Model to `Women' and `Life'}\label{appendix:role-dist}
Distribution of MF Roles Assigned by the Model to `Women' and `Life' when mentioned by Democrats and Republicans, respectively, are shown in Table \ref{tab:role-dist-women} and Table \ref{tab:role-dist-life}, respectively.

\begin{table*}[h!]
\begin{center}
 \scalebox{0.60}{\begin{tabular}{>{\arraybackslash}m{12cm}|>{\arraybackslash}m{12cm}} 
 \hline 
\textbf{(Predicted MF) Tweet} & \textbf{Comment} \\ [0.5ex]
 \hline
 \textbf{(CARE/HARM)} The U.S. Senate is set to vote on commonsense legislation to protect \textcolor{purple}{$[$}unborn babies\textcolor{purple}{$]_{TARGET-OF-CARE}$} who can feel pain. Retweet if you Stand For \textcolor{purple}{$[$}Life\textcolor{purple}{$]_{PROVIDING-CARE}$}! &  MF prediction is `Care/Harm', possibly because there is a notion of protecting babies. In MF role prediction, the model makes mistake when there are multiple mention of the same entity, possibly because of constraint $c_2$ but still assigns a positive role to `Life', possibly because of constraint $c_3$.\\
 \hline
 \textbf{(LOYALTY/BETRAYAL)} I will always, always, ALWAYS be proud to Stand 4 \textcolor{purple}{$[$}Life\textcolor{purple}{$]_{BEING-LOYAL}$}. I'm so grateful to @TXRightToLife for their support and pledge to never stop fighting for the \textcolor{purple}{$[$}unborn\textcolor{purple}{$]_{TARGET-OF-LOYALTY}$}. Now, Texas, let's get out and vote to \#KeepTexasRed! & MF prediction is correct. In MF role prediction, the model makes mistake when there are multiple mention of the same entity, possibly because of constraint $c_2$ but still assigns a positive role to `Life', possibly because of constraint $c_3$.\\
 \hline
 \textbf{(PURITY/DEGRADATION)} \textcolor{purple}{$[$}Planned Parenthood\textcolor{purple}{$]_{VIOLATING-PURITY}$} is suing our state to expand their abortion-on-demand agenda. RT if you stand for \textcolor{purple}{$[$}life\textcolor{purple}{$]_{PRESERVING-PURITY}$}! & MF prediction is wrong. Still a positive role is assigned to `life' and a negative role is assigned to `Planned Parenthood', possibly because of constraint $c_3$.\\
 \hline
 \end{tabular}}
\caption{Qualitative error analysis for MF role prediction for the entity `Life' in Republicans.}
\label{tab:qualitative-error-analysis-life}
\end{center}
\end{table*}

\begin{table}[h!]
\begin{center}
\scalebox{0.6}{\begin{tabular}{>{\arraybackslash}m{5cm}>{\centering\arraybackslash}m{3cm}} 
 \toprule
 \textbf{\textsc{Moral Roles}} & \textbf{\textsc{\% of time Assigned by the Model}}\\
 \midrule
    Target of fairness/cheating & 0.624\\
    Target of care/harm & 0.216\\
    Failing authority over & 0.076\\
    Target of loyalty/betrayal & 0.075\\
    Target of purity/degradation & 0.006\\
    Entity providing care & 0.001\\
    Entity being loyal & 0.001\\
 \bottomrule
 \end{tabular}}
\caption{Distribution of MF roles assigned to `Women' by the model when mentioned by the `Democrats'.}
\label{tab:role-dist-women}

 \scalebox{0.6}{\begin{tabular}{>{\arraybackslash}m{5cm}>{\centering\arraybackslash}m{3cm}} 
 \toprule
 \textbf{\textsc{Moral Roles}} & \textbf{\textsc{\% of time Assigned by the Model}}\\
 \midrule
    Target of purity/degradation & 0.501\\
    Target of care/harm & 0.266\\
    Target of loyalty/betrayal & 0.151\\
    Target of fairness/cheating & 0.038\\
    Failing authority over & 0.018\\
    Entity being loyal & 0.008\\
    Entity providing care & 0.005\\
    Entity preserving purity & 0.004\\
    Justified authority & 0.004\\
    Entity ensuring fairness & 0.003\\
    Justified authority over & 0.002\\
 \bottomrule
 \end{tabular}}
    
\caption{Distribution of MF roles assigned to `Life' by the model when mentioned by the `Republicans'.}
\label{tab:role-dist-life}
\end{center}
\end{table}

\begin{table}[h!]
\begin{center}
\scalebox{0.60}{
\begin{tabular}{>{\arraybackslash}m{4cm}|>{\centering\arraybackslash}m{2cm}|>{\centering\arraybackslash}m{2cm}||>{\centering\arraybackslash}m{2.3cm}}
    \hline
    \textbf{\textsc{Moral Foundations}} & \textbf{\textsc{\%Used in Left}} & \textbf{\textsc{\%Used in Right}} & \textbf{\textsc{\%Predicted in Total}}\\ [0.5ex]
    \hline
    \textsc{\textbf{Care/harm}}             &   0.23    &   0.24    &   0.24\\  
    \textsc{\textbf{Fairness/Cheating}}     &   0.52    &   0.14    &   0.35\\
    \textsc{\textbf{Loyalty/Betrayal}}      &   0.09    &   0.17    &   0.13\\
    \textsc{\textbf{Authority/Subv.}}       &   0.13    &   0.12    &   0.13\\
    \textsc{\textbf{Purity/Degrad.}}    &   0.02    &   0.34    &   0.17\\
    \hline
\end{tabular}
}
\scalebox{0.5}{\begin{tabular}{>{\arraybackslash}m{3.5cm}|>{\centering\arraybackslash}m{1cm}|>{\centering\arraybackslash}m{1cm}||>{\centering\arraybackslash}m{1cm}|>{\centering\arraybackslash}m{1cm}||>{\centering\arraybackslash}m{3cm}} 
 \hline 
\textbf{\textsc{Morals}} & \multicolumn{2}{c}{\textsc{\textbf{\%Used in Left}}} & \multicolumn{2}{c}{\textsc{\textbf{\%Used in Right}}} & \multicolumn{1}{c}{\textsc{\textbf{\%Pred. in Total}}}\\ [0.5ex]
\cline{2-6}
&   \textsc{Pre}   &   \textsc{Post}   &   \textsc{Pre}   &   \textsc{Post} & \textsc{Pre+Post} \\
 \hline
 \textsc{\textbf{Care/Harm}}             & 0.33  & 0.4  & 0.27 & 0.42 & 0.37  \\
 \textsc{\textbf{Fairness/Cheating}}     & 0.05  & 0.02  & 0.04 & 0.03 & 0.03    \\
 \textsc{\textbf{Loyalty/Betrayal}}      & 0.19  & 0.15  & 0.22 & 0.25 & 0.19   \\
 \textsc{\textbf{Authority/Subv.}}      & 0.42  & 0.42  & 0.46 & 0.29 & 0.40   \\
 \textsc{\textbf{Purity/Degrad.}}       & 0.01  & 0.01  & 0.02 & 0.01 & 0.1   \\
 \hline
 \end{tabular}}
\caption{Moral foundation usage by ideologies on topic Abortion (top); and pre and post the US Capitol incident on Jan 6, 2021 (bottom). The percentage of time each moral foundation is predicted by the model are shown in the right-most column of each table.}
\label{tab:mf-usage-abortion-capitol-hill}
\end{center}
\end{table}

\subsection{Qualitative Evaluation of MF Role Prediction for `Life' in Republicans}\label{appendix:life-qualitative-eval}
Some tweets mentioning `Life' by the Republicans and the predicted MF and MF roles are shown in Table \ref{tab:qualitative-error-analysis-life}.


\section{Analysis of Political Discussion}

\subsection{Moral Foundation Usage}\label{appendix:moral-foundation-usage}
Wikipidia link to the US Capitol incident: \href{https://en.wikipedia.org/wiki/2021_storming_of_the_United_States_Capitol}{Link}. 
The distribution of the usage of different moral foundations on the topics Abortion and US Capitol event can be found in Table \ref{tab:mf-usage-abortion-capitol-hill} (top) and Table \ref{tab:mf-usage-abortion-capitol-hill} (bottom), respectively.



\subsection{Most Targeted Entities and Entity Relationship Graphs After the Event US Capitol Storming (2021)}\label{appendix:entity-relation-ch}
\begin{figure}[h!]
    \centering
    \includegraphics[width=0.4\textwidth]{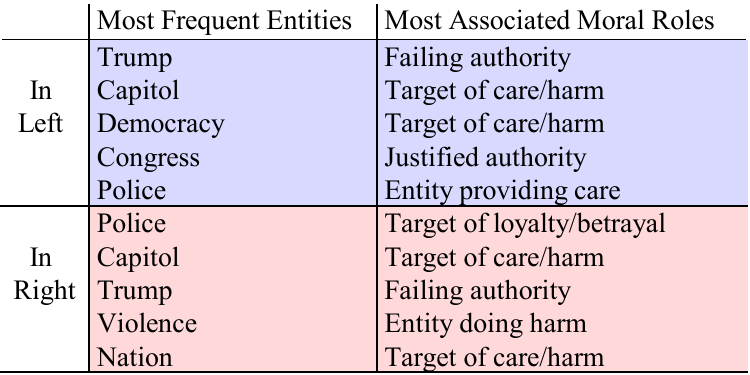}
    \caption{Most frequent entities \& most associated MF roles after the event US Capitol Storming (2021).}
    \label{fig:most_used_entities_ch}
    \begin{subfigure}[t]{0.35\textwidth}
        \centering
        \includegraphics[width=1\textwidth]{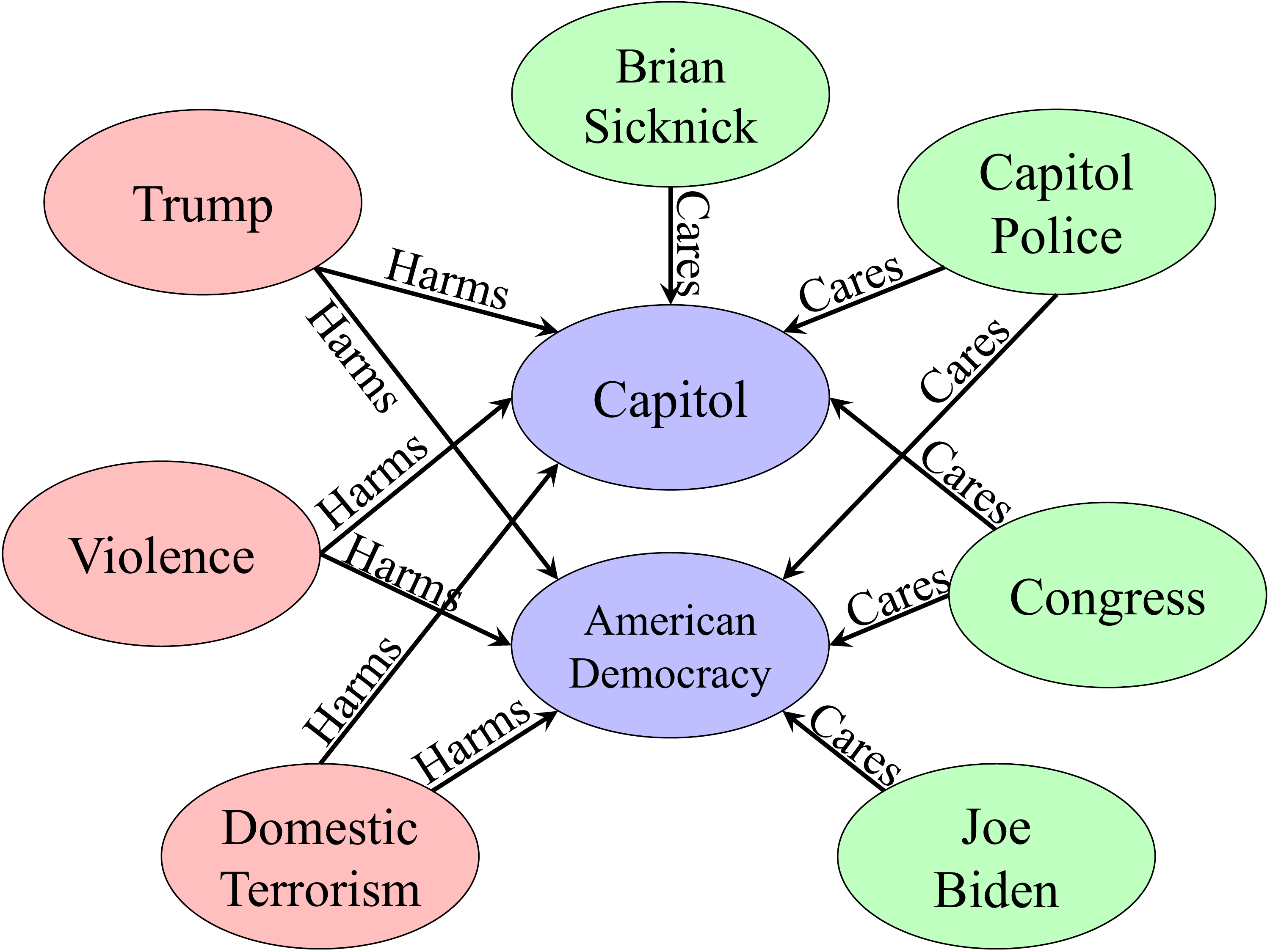}
        \caption{Entity relation in left}
        \label{fig:entity-relation-left}
    \end{subfigure}%
    \\
    \begin{subfigure}[t]{0.35\textwidth}
        \centering
        \includegraphics[width=1\textwidth]{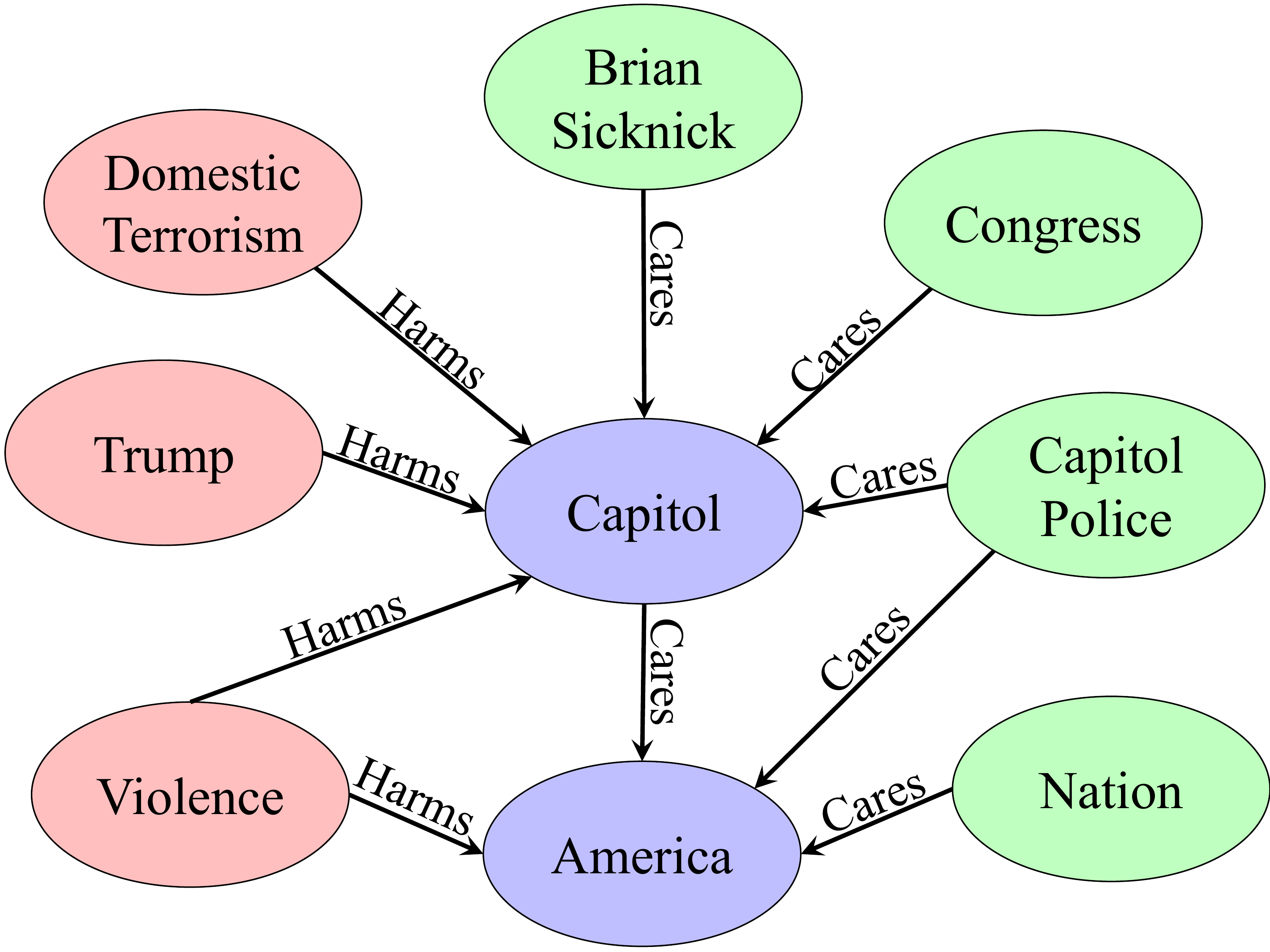}
        \caption{Entity relation in right}
        \label{fig:entity-relation-right}
    \end{subfigure}
    \caption{Entity relationship graphs for \textbf{Care/Harm} after the US Capitol Storming (2021).}
    \label{fig:entity-relation-ch}
\end{figure}

The most targeted entities and entity relationship graphs after the US Capitol Storming (2021) are shown in Figure \ref{fig:most_used_entities_ch} and \ref{fig:entity-relation-ch}, respectively.

